\documentclass{m3ailpreprint}

\usepackage{amsmath,amsfonts,bm}

\def\eqref#1{equation~\ref{#1}}

\def\1{\bm{1}}

\DeclareMathAlphabet{\mathsfit}{\encodingdefault}{\sfdefault}{m}{sl}
\SetMathAlphabet{\mathsfit}{bold}{\encodingdefault}{\sfdefault}{bx}{n}

\usepackage{url}
\usepackage[utf8]{inputenc} 

\usepackage{threeparttable}
\usepackage{array}
\usepackage{tabularx}
\usepackage{adjustbox}
\usepackage{float}
\usepackage{wrapfig}
\usepackage{colortbl}

\usepackage{nicefrac}       
\usepackage{textcomp}
\usepackage{verbatim}

\usepackage{soul}      

\definecolor{myhighlight}{RGB}{255,255,153}
\definecolor{tableOnline}{HTML}{FFF3D6}
\definecolor{tableOffline}{HTML}{F6E8F2}
\definecolor{tableOurs}{HTML}{E1EFF7}

\sethlcolor{myhighlight}

\tcbuselibrary{breakable,listings}
\usepackage{enumitem}


\makeatletter
\newif\if@restonecol
\makeatother

\usepackage[linesnumbered,ruled,vlined]{algorithm2e}
\usepackage{algpseudocode}

\SetKwComment{Comment}{$\triangleright$\ }{}
\SetKwRepeat{Do}{do}{while}

\definecolor{b_proj}{HTML}{D71D3D}
\definecolor{b_mse}{HTML}{FFD966}
\definecolor{b_act}{HTML}{999999}

\renewcommand{\paragraph}[1]{\textbf{#1}\ }

\title{ShallowStream: Index Shallow then Answer Deep for Streaming Video Understanding}

\author[1,*]{Jitai Hao}
\author[1,*]{Ke Yang}
\author[1]{Di Yan}
\author[2]{Fan Liu}
\author[1,\dagger]{Qiang Huang}
\author[1,\dagger]{Jun Yu}

\affiliation[1]{Harbin Institute of Technology (Shenzhen)}
\affiliation[2]{Southeast University}

\contribution{$^*$Equal contribution.}
\contribution{$^\dagger$Corresponding authors.}

\abstract{%
Streaming video understanding is essential for real-time embodied and assistive systems, but processing open-ended streams with multimodal large language models (MLLMs) is computationally expensive.
Existing methods reduce visual tokens or stored context, yet still apply full-depth prefill to admitted frames, causing repeated computation and depth-proportional KV-cache growth.
We observe that shallow MLLM layers already provide effective signals for retrieving question-relevant evidence.
Building on this observation, we propose \textbf{ShallowStream}, which decouples lightweight stream processing from selective full-depth answering.
During streaming, ShallowStream processes frames only through shallow layers, using their KVs as a historical index with optional cluster compression.
At query time, a text-only gate activates cross-layer token voting and diversity-aware retrieval when needed, after which only the selected history and recent context undergo full-depth prefill.
ShallowStream reduces per-frame prefill and 20-second end-to-end latency by up to \textbf{52.1$\times$} and \textbf{15.3$\times$}, respectively, while improving average scores by around \textbf{1.7 points} over strong baselines.

}

\checkdata[Correspondence]{huangqiang@hit.edu.cn, yujun@hit.edu.cn}
\checkdata[Repository]{\url{https://github.com/CURRENTF/ShallowStream}}

\begin{document}
\maketitle

\section{Introduction}
\label{sect:intro}

\begin{wrapfigure}[13]{r}{0.6\textwidth}
  \vspace{-2.0em}
  \centering
  \includegraphics[width=0.90\linewidth]{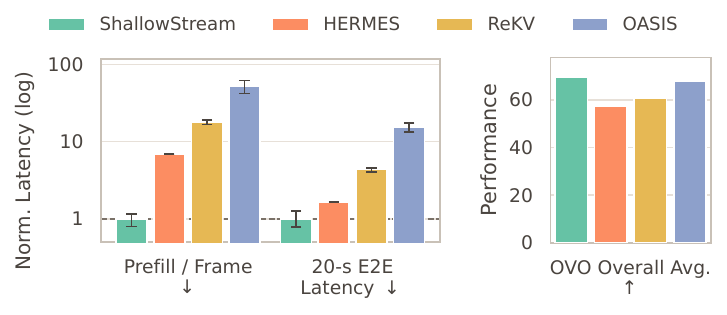}
  \captionsetup{skip=0.35em}
  \caption{\textbf{Continuous-stream efficiency.} Per-frame prefill and end-to-end latency with one query every 20s. Latencies and their 95\% confidence intervals are scaled by the corresponding ShallowStream mean and shown on a logarithmic scale. Performance is the OVO-Bench average.}
  \label{fig:latency-performance-teaser}
\end{wrapfigure}

Multimodal Large Language Models (MLLMs) have demonstrated strong capabilities across diverse visual tasks~\citep{Bai2025Qwen25VLTR, Chen2023InternVS, Bai2025Qwen3VLTR}.
These advances have spurred growing interest in streaming video understanding, which supports applications such as embodied intelligence, autonomous driving, surveillance and early warning, AR glasses, sports commentary, and fitness coaching~\citep{Chen2024VideoLLMonlineOV, Li2025OVOBenchHF, Lin2024StreamingBenchAT, Di2025StreamingVQ}.
Unlike offline long-video understanding, where the complete video and question are available before inference, streaming video understanding must causally process an open-ended stream without access to future frames or advance knowledge of future questions.

Streaming workloads are inherently \emph{asymmetric}: frames arrive continuously, whereas queries occur only intermittently \citep{Qian2024StreamingLV, Wang2024VideoLLaMBLS, Chen2025StreamKVSV, Yang2025StreamMemQK, Chen2025StreamingTOMST, Lu2026VistaSO, Ning2025LiveVLMEO, Kim2025VRexRS, Zhang2026WeaveTimeSF}.
During \emph{query-agnostic stream processing}, a model encodes incoming frames and maintains historical context before future questions are known \citep{Di2025StreamingVQ, Zhang2026HERMESKC}.
During \emph{query-time answering}, it determines whether earlier context is needed, retrieves relevant evidence, and generates an answer \citep{Shen2026ASB, Liang2026OASISOH}.
Because stream processing is continuous while question answering is sparse, per-frame prefill becomes a \textbf{first-order system cost}.
As shown in Figure~\ref{fig:latency-performance-teaser}, the substantial per-frame overhead of representative methods accumulates into high end-to-end latency.

This setting exposes three limitations in existing streaming systems:
\begin{itemize}[nolistsep, left=1pt]
  \item \textbf{Expensive Stream Processing (\emph{Stream Processing}).}
  Many KV-centric methods prefill every incoming frame through the full Transformer stack before knowing whether its deep representations will ever be used~\citep{Di2025StreamingVQ, Kim2025InfiniPotVMK}.
  Subsequent cache reduction lowers memory consumption but cannot recover the computation already spent constructing deep KVs.

  \item \textbf{Lossy Evidence Reduction (\emph{Stream Processing}).}
  Query-agnostic compression, merging, or eviction limits KV-cache growth but may discard evidence needed by future retrospective queries~\citep{Chen2025StreamKVSV, Yang2025StreamMemQK, Li2024SCBenchAK}; retaining complete states preserves evidence but incurs prohibitive memory costs and bandwidth-limited offloading \citep{Kim2025VRexRS}.

  \item \textbf{Inefficient History Use (\emph{Query-Time Answering}).}
  Supplying history to every query increases latency and can interfere with current-scene perception~\citep{Shen2026ASB}.
  Conversely, on-demand methods may first perform costly answer-level inference to determine whether historical evidence is needed~\citep{Liang2026OASISOH, Zhang2026WeaveTimeSF}.
\end{itemize}

\vspace{-0.25em}
Our key observation is that historical evidence can be retrieved effectively without processing every frame through the full model depth.
As shown in Figure~\ref{fig:layer_retrieval_ability}, strong retrieval capability already emerges at layer 4 of 28 in Qwen3-VL-8B~\citep{Bai2025Qwen3VLTR} and layer 3 of 32 in LLaVA-OneVision-7B~\citep{Li2024LLaVAOneVisionEV}.
Deeper stream-time processing therefore incurs substantial additional cost while providing limited benefit for evidence selection.

Motivated by this observation, we propose \textbf{ShallowStream}, which addresses these limitations by decoupling lightweight, query-agnostic stream processing from selective full-depth answering:
\begin{itemize}[nolistsep, left=1pt]
  \item \textbf{Shallow Encoding:}
  During continuous stream processing, ShallowStream propagates each incoming video unit through only the shallow layers of the language Transformer, avoiding unnecessary deep-layer prefill and substantially reducing steady-state computation.

  \item \textbf{Full-History Visual Indexing:}
  The resulting shallow-layer KVs form a lightweight index over the observed stream, preserving access to historical evidence without retaining full-depth states for every frame.
  Optional long-cluster compression further bounds memory over long streams.

  \item \textbf{Selective Query-Time Answering:}
  A lightweight query-logit gate activates retrieval when needed, after which shallow-layer token voting and diversity-aware selection identify complementary evidence for full-depth answering with recent context.
\end{itemize}

\vspace{-0.25em}
Extensive experiments show that, without task-specific training, ShallowStream reduces per-frame prefill and 20-second end-to-end latency on long videos by up to \textbf{52.1$\times$} and \textbf{15.3$\times$}, respectively, while improving average scores by approximately \textbf{1.7 points} over strong existing baselines.
These results show that shallow indexing with selective deep answering provides an effective and efficient operating point for continuous video understanding.

\section{Related Work}
\label{sect:related}

Unlike offline long-video understanding, streaming video understanding causally processes sequential frames without access to future content or advance knowledge of questions, which may arrive at any time \citep{Chen2024VideoLLMonlineOV, Qian2024StreamingLV, Lin2024StreamingBenchAT, Yang2025SVBenchAB, Wang2024VideoLLaMBLS, Li2025OVOBenchHF, Di2025StreamingVQ, Yang2025StreamMemQK, Ning2025LiveVLMEO, Shen2026ASB}.
Its asymmetric workload has two stages: query-agnostic stream processing and query-time answering.
We review prior work along these two stages.

\paragraph{Query-Agnostic Stream Processing.}
Before questions arrive, existing methods represent the stream using recurrent memory, KV-caches, or compact visual representations.
VideoStreaming~\citep{Qian2024StreamingLV} propagates memory across clips, while VideoLLaMB~\citep{Wang2024VideoLLaMBLS} links them through recurrent memory bridges.
KV-centric methods maintain layer-wise visual states:
ReKV~\citep{Di2025StreamingVQ} offloads complete video KVs to host memory or disk; StreamKV~\citep{Chen2025StreamKVSV} applies segment-wise adaptive compression; and StreamMem~\citep{Yang2025StreamMemQK}, InfiniPot-V~\citep{Kim2025InfiniPotVMK}, and HERMES~\citep{Zhang2026HERMESKC} bound cache growth using proxy-query attention, temporal redundancy with Value Norm, and hierarchical recency-attention cues, respectively.
Other methods reduce visual input before MLLM prefill:
TimeChat-Online~\citep{Yao2025TimeChatOnline8V} and StreamingTOM~\citep{Chen2025StreamingTOMST} compress visual tokens, whereas Vista~\citep{Lu2026VistaSO} indexes compact scene representations and offloads high-resolution frames to CPU memory.

Despite these differences, every frame or token admitted to the MLLM still traverses its \emph{full} Transformer depth.
Thus, these methods reduce the content processed or retained but leave model depth, and its associated stream-time cost, unchanged.
ShallowStream instead reduces computation at its source: each frame traverses only the shallow layers, whose KVs also form a lightweight full-history index, while optional cluster compression bounds memory over long streams.


\paragraph{Query-Time Answering.}
Existing methods differ in \emph{how} and \emph{when} they access history.
StreamMem~\citep{Yang2025StreamMemQK}, InfiniPot-V~\citep{Kim2025InfiniPotVMK}, and HERMES~\citep{Zhang2026HERMESKC} answer from compressed KVs, whereas ReKV~\citep{Di2025StreamingVQ}, StreamKV~\citep{Chen2025StreamKVSV}, LiveVLM~\citep{Ning2025LiveVLMEO}, and V-Rex~\citep{Kim2025VRexRS} retrieve historical states on demand.
StreamingTOM~\citep{Chen2025StreamingTOMST} retrieves token groups from quantized memory, while Vista~\citep{Lu2026VistaSO} recalls compact scene representations with high-resolution content.
Regarding \emph{when}, SimpleStream~\citep{Shen2026ASB} shows that unnecessary history increases latency and can impair current-scene perception.
OASIS~\citep{Liang2026OASISOH} and WeaveTime~\citep{Zhang2026WeaveTimeSF} therefore retrieve history only when an initial answer lacks support or has high uncertainty, but this requires answer-level multimodal inference followed by another inference when retrieval is activated.

ShallowStream separates routing from answering: a single text-only query-logit evaluation determines whether history is needed without generating a candidate answer.
If activated, cross-layer token voting and diversity-aware selection retrieve relevant, nonredundant units, and only the selected history and recent context undergo full-depth visual prefill.
More related work is provided in Appendix \ref{appendix:other_related_work}.

\section{Observations}
\label{sect:observations}

\paragraph{Background and Notation.}
A streaming video arrives causally as a sequence of video units $\mathcal{X} = \{x_t\}_{t \in \{1, 2, \cdots\}}$, where $x_t$ is the $t$-th unit.
Each unit contains two sampled frames for Qwen3-VL and one for LLaVA-OneVision.
At query time $T$, the system has access only to the observed prefix $\mathcal{X}_{\leq T}=\{x_t\}_{t \in \{1, \cdots, T\}}$; neither the future stream $\mathcal{X}_{>T}$ nor the incoming question $q_T$ is known beforehand.
When unambiguous, we write $q=q_T$ and denote the generated answer by $\bm{y}$.

We build on a pretrained MLLM comprising a vision encoder $E_{\mathrm{v}}$ and an $L$-layer language Transformer $\mathcal{F}=\{F^i\}_{i=0}^{L-1}$.
The vision encoder maps each unit $x_t$ to visual tokens indexed by $\mathcal{P}_t$.
Each Transformer layer $F^i$ maps hidden states $\bm{H}_t^i$ to $\bm{H}_t^{i+1}$ while producing key and value states $\bm{K}_t^i$ and $\bm{V}_t^i$.
We define the pruning boundary $P\ll L$ as the first deep layer: layers $[0,P)$ constitute the \textbf{shallow MLLM}, layers $[P,L)$ constitute the \textbf{deep MLLM}, and together they form the full MLLM.

\begin{figure*}[t]
  \centering
  \begin{minipage}[t]{0.49\textwidth}
    \centering
    \includegraphics[width=0.99\linewidth]{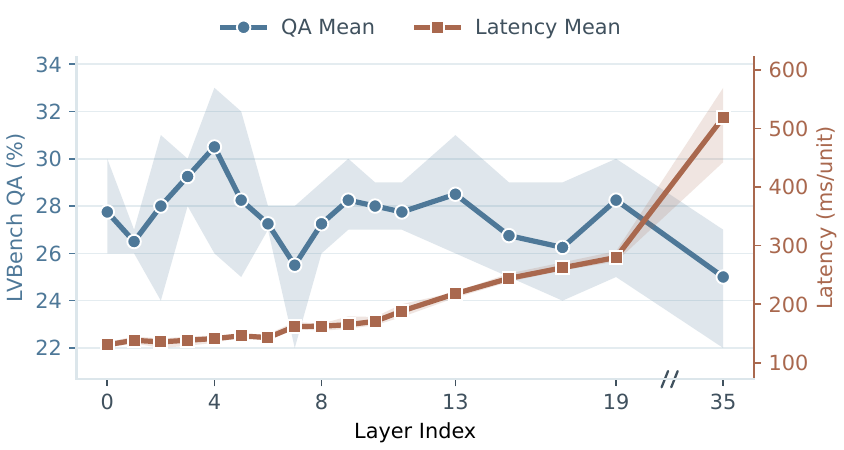}
  \end{minipage}
  \hspace{0.01\textwidth}%
  \begin{minipage}[t]{0.49\textwidth}
    \centering
    \includegraphics[width=0.99\linewidth]{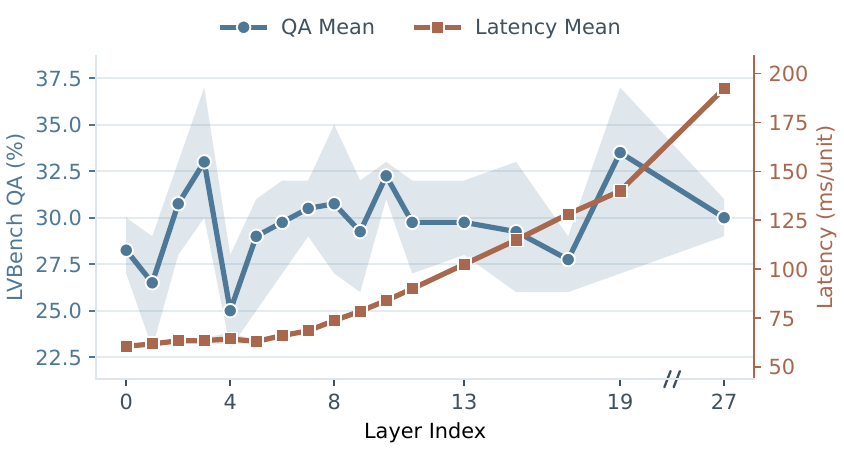}
  \end{minipage}
  \caption{\textbf{Layer-wise retrieval quality and stream-time prefill cost on LVBench.}
  Results for Qwen3-VL-8B (\emph{left}) and LLaVA-OneVision-7B (\emph{right}) as the retrieval layer varies, with all other settings fixed. QA measures full-depth answer accuracy using units selected at that layer, while latency measures stream-time prefill per video unit up to that layer.}
  \label{fig:layer_retrieval_ability}
  \vspace{-0.5em}
\end{figure*}

\paragraph{Observation: Shallow Layers Suffice for Retrieval.}
We find that shallow MLLM representations effectively retrieve question-relevant units from long histories, consistent with evidence that shallow and middle layers preserve rich semantic information~\citep{Xin2020EarlyEB, Meng2022LocatingAE, Hao2025OmniKVDC}.
To isolate retrieval quality from answer generation, we conduct a layer-wise LVBench diagnostic~\citep{Wang2024LVBenchAE} under ReKV's attention-sink, bounded-window prefill setting~\citep{Di2025StreamingVQ}: each candidate layer ranks historical units, and the full-depth model answers using only the highest-ranked units.
Appendix~\ref{appendix:sliding_window_retrieval_depth} provides a dense full-history control.

Figure~\ref{fig:layer_retrieval_ability} shows that effective retrieval signals emerge in shallow layers, while stream-time prefill cost continues to increase with depth.
This quality-cost separation motivates ShallowStream: it uses only shallow layers for continuous video encoding and evidence indexing, deferring full-depth computation until a question arrives and applying it only to the retrieved evidence.

\section{ShallowStream}
\label{sect:method}

\begin{figure*}[t]
\centering
\includegraphics[width=0.99\textwidth]{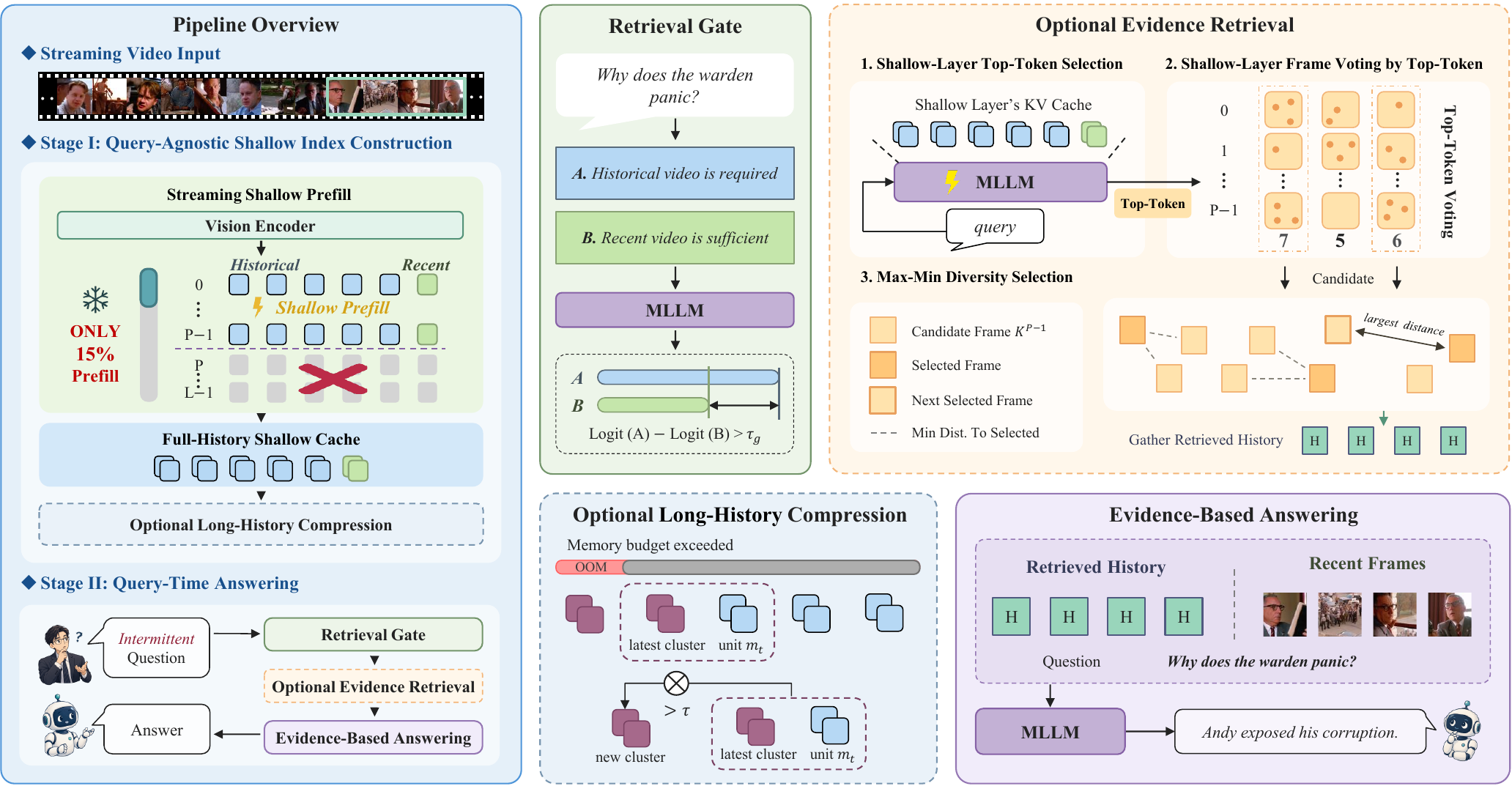}
\caption{\textbf{Overview of ShallowStream.}
\textbf{Stage I: Query-Agnostic Shallow Index Construction} (Section~\ref{sect:method:stream_encoding}). Incoming video units traverse only the shallow MLLM layers, whose KVs form a lightweight index that can optionally be compressed into fixed-size historical clusters.
\textbf{Stage II: Query-Time Answering} (Section~\ref{sect:method:query_answering}).
A text-only query-logit gate determines whether to retrieve history; if activated, shallow-layer attention, cross-layer unit voting, and max-min diversity selection identify complementary evidence for full-depth answering alongside recent context.}
\label{fig:method_overview}
\end{figure*}

We present ShallowStream, which assigns different computation budgets to the two asymmetric stages of streaming video understanding.
Figure \ref{fig:method_overview} illustrates the pipeline.

\subsection{Query-Agnostic Shallow Index Construction}
\label{sect:method:stream_encoding}

\paragraph{Streaming Shallow Prefill.}
Each incoming video unit $x_t$ is first encoded as $\bm{H}_t^0=E_{\mathrm{v}}(x_t)$ and then propagated only through the shallow layers $[0,P)$:
\begin{equation}
\label{eqn:shallow_prefill}
  (\bm{H}_t^{i+1},\bm{K}_t^i,\bm{V}_t^i) = F^i(\bm{H}_t^i; \mathcal{C}_{t-1}^i),
  \qquad i \in \{0,\cdots,P-1\},
\end{equation}
where $\mathcal{C}_{t-1}^i$ is the causal streaming KV cache available to layer $i$.
The detailed memory retains $\bm{H}_t^0$ and the resulting shallow KVs, allowing a selected unit to be re-prefilled later from layer $0$ without storing any deep-layer KVs.
Under optional long-cluster compression, older per-unit states are replaced by the fixed-size representatives defined below.
By skipping the deep layers $\{F^P,\cdots,F^{L-1}\}$ during streaming, ShallowStream reduces per-unit Transformer computation from $L$ layers to $P$.

\paragraph{Full-History Shallow Cache.}
To help queries access distant history, ShallowStream retains the KVs of all observed units at each shallow layer:
\begin{equation}
\label{eqn:shallow_streaming_cache}
  \mathcal{C}_T^{\mathrm{shallow}} =
  \{
    (\bm{K}_{1:T}^{i},\bm{V}_{1:T}^{i})
  \}_{i=0}^{P-1}.
\end{equation}
Here, $\bm{K}_{1:T}^{i}$ and $\bm{V}_{1:T}^{i}$ concatenate the KVs of $\{x_1,\cdots,x_T\}$ at layer $i$.
For each detailed unit $x_t$, we store its visual encoding, shallow-layer KV cache, and timestamp as $m_t=\{\bm{H}_t^0,(\bm{K}_t^i,\bm{V}_t^i)_{i=0}^{P-1},t\}$.
The cached KVs allow an incoming query to attend to the observed history throughout the shallow layers $[0,P)$, while the visual-token keys support fine-grained evidence retrieval.

For memory management and diversity-aware selection, we summarize each unit with a fixed-length descriptor.
Specifically, we average its raw visual keys from the final shallow layer over the visual-token indices $\mathcal{P}_t$ and apply $\ell_2$ normalization:
\begin{equation}
\label{eqn:memory_key}
  \bm{k}_t = \operatorname{Norm}\Big(
    \frac{1}{|\mathcal{P}_t|}
    \sum_{p\in\mathcal{P}_t}
    \operatorname{vec}(\bm{K}_{t,p}^{P-1})
  \Big),
\end{equation}
where $\operatorname{vec}(\cdot)$ flattens all key heads and $\operatorname{Norm}(\bm{z})=\bm{z}/\|\bm{z}\|_2$.
The descriptor $\bm{k}_t$ captures unit-level structure, while the unpooled shallow keys retain token-level retrieval signals.

At query time $T$, we divide the streaming memory $\mathcal{M}_T$ into recent and historical contexts.
The latest $N_r$ units form $\mathcal{R}_T$, and all earlier units form $\mathcal{H}_T$:
\begin{equation}
\label{eqn:context_partition}
  \mathcal{M}_T
  =\mathcal{H}_T\cup\mathcal{R}_T,
  \qquad
  \mathcal{R}_T=\{m_t\mid T-N_r<t\leq T\},
  \quad
  \mathcal{H}_T=\mathcal{M}_T\setminus\mathcal{R}_T.
\end{equation}
Both sets retain KVs at every shallow layer.
The recent context is always available for current-scene perception, whereas historical evidence is retrieved only when required.

\paragraph{Optional Long-History Compression.}
When the memory footprint of $\mathcal{M}_T$ exceeds a budget $B$, ShallowStream compresses older history into long-term clusters $\mathcal{L}_T$, while preserving the recent context $\mathcal{R}_T$ and near-term history $\mathcal{S}_T \subset \mathcal{H}_T$ in detail:
\begin{equation}
\label{eqn:optional_context_compression}
  \mathcal{M}_T^{(B)}=
  \begin{cases}
    \mathcal{H}_T\cup\mathcal{R}_T,
      & \operatorname{Mem}(\mathcal{M}_T)\leq B,\\
    \mathcal{L}_T\cup\mathcal{S}_T\cup\mathcal{R}_T,
      & \operatorname{Mem}(\mathcal{M}_T)>B.
  \end{cases}
\end{equation}
Clusters are constructed online over temporally adjacent units.
When a unit $m_t$ leaves the detailed window, it joins the latest cluster $C_j$ if its descriptor $\bm{k}_t$ is sufficiently similar to the cluster centroid $\bm{c}_j$; otherwise, it starts a new cluster:
\begin{equation}
\label{eqn:optional_long_memory_clustering}
  C(m_t)=
  \begin{cases}
    C_j, & \bm{k}_t^{\top}\bm{c}_j\geq\tau,\\
    C_{j+1}, & \text{otherwise},
  \end{cases}
  \qquad
  \bm{c}_j\leftarrow
  \operatorname{Norm}\!\left(
    \frac{n_j\bm{c}_j+\bm{k}_t}{n_j+1}
  \right),
\end{equation}
where $n_j$ is the current size of $C_j$. Its representative shallow KVs are updated by a running average:
\begin{equation}
\label{eqn:optional_long_memory_representative}
  (\overline{\bm{K}}_j^i,\overline{\bm{V}}_j^i)
  \leftarrow
  \frac{n_j(\overline{\bm{K}}_j^i,\overline{\bm{V}}_j^i)
  +(\bm{K}_t^i,\bm{V}_t^i)}{n_j+1}.
\end{equation}

\vspace{-1.0em}
Each cluster replaces its members' archived states with one input-level visual representative $\overline{\bm{H}}_j^0$, obtained from the source unit nearest $\bm{c}_j$ or by averaging visual embeddings.
If selected, only this representative is transferred, yielding fixed per-cluster storage and cluster-size-independent host-to-device transfer.
Compression is disabled when the uncompressed $\mathcal{H}_T\cup\mathcal{R}_T$ fits in memory.

\subsection{Query-Time Answering}
\label{sect:method:query_answering}

\paragraph{Retrieval Gate.}
Retrospective questions require historical evidence, whereas current-scene questions can often be answered from $\mathcal{R}_T$ alone.
ShallowStream distinguishes them without generating a candidate answer by placing $q$ in a fixed few-shot routing prompt $r(q)$ with two single-token choices: $A$ for historical retrieval and $B$ for recent context only.
A single text-only forward pass produces:
\begin{equation}
\label{eqn:query_gate_score}
  \ell_{\mathrm{ret}}(q)
  =\operatorname{Logit}_{A}(r(q)),
  \qquad
  \ell_{\mathrm{recent}}(q)
  =\operatorname{Logit}_{B}(r(q)).
\end{equation}
Retrieval is activated when their difference exceeds a backbone-specific threshold $\tau_g$:
\begin{equation}
\label{eqn:query_gate}
  g(q) = \mathbf{1}[
    \ell_{\mathrm{ret}}(q)-\ell_{\mathrm{recent}}(q) \geq \tau_g
  ].
\end{equation}
For $g(q)=0$, the model uses only $\mathcal{R}_T$; for $g(q)=1$, it retrieves from detailed history and optional long-term clusters.
We calibrate $\tau_g$ once on a benchmark-independent set and freeze it for evaluation, requiring neither a separately trained router nor benchmark labels (Appendix~\ref{appendix:query_gate_calibration}).

\paragraph{Optional Evidence Retrieval.}
When retrieval is activated, we append $q$ to the retained shallow history and propagate it through layers $[0,P)$:
\begin{equation}
\label{eqn:query_full_history_attention}
  (\bm{H}_q^{i+1},\bm{Q}_q^i)
  =F^i(\bm{H}_q^i;\mathcal{C}_T^i),
  \qquad i \in \{0,\cdots,P-1\},
\end{equation}
where $\mathcal{C}_T^i$ is the retained layer-$i$ KV cache associated with $\mathcal{M}_T$, or $\mathcal{M}_T^{(B)}$ under Equation~\ref{eqn:optional_context_compression}, and $\bm{Q}_q^i$ denotes the query states.
The query attends to exact full-history shallow KVs without compression, or to detailed recent and representative historical KVs with compression.

For detailed history, ShallowStream retains token-level retrieval signals.
Let $\mathcal{P}(\mathcal{S}_T)=\bigcup_{m_t\in\mathcal{S}_T}\mathcal{P}_t$ be the candidate visual tokens, with $\mathcal{S}_T=\mathcal{H}_T$ when uncompressed.
At each shallow layer $i$, we compute scaled, RoPE-aware Q-K attention from the final prompt token to all candidates and average over attention heads.
If $\alpha_{i,p}$ denotes this score, we retain the top $M$ tokens at each layer:
\begin{equation}
\label{eqn:token_vote_candidates}
  \mathcal{V}_i^{M}
  =\operatorname{TopM}_{p\in\mathcal{P}(\mathcal{S}_T)}\alpha_{i,p},
  \qquad i \in \{0,\cdots,P-1\}.
\end{equation}
Each unit $m_t\in\mathcal{S}_T$ receives one vote per selected token: $v_t= \sum_{i=0}^{P-1}\sum_{p\in\mathcal{P}_t} \mathbf{1}\![p\in\mathcal{V}_i^{M}]$.

We retain the top $4K$ units ranked by $v_t$, breaking ties by attention mass, then use max-min diversity selection~\citep{Alvar2025DivPruneDV} in the descriptor space of Equation~\ref{eqn:memory_key} to choose $K$ nonredundant units.
Each selected unit is then expanded with its temporal neighbors to restore local context:
\begin{equation}
\label{eqn:selected_evidence}
  \mathcal{E}_{\mathrm{det}}(q)=
  \begin{cases}
    \mathcal{R}_T,
      & g(q)=0,\\
    \mathcal{R}_T
    \cup\operatorname{Nbr}\!(\operatorname{DivTopK}_{K}(\mathcal{S}_T;v,\bm{k})),
      & g(q)=1,
  \end{cases}
\end{equation}
Here, $\operatorname{DivTopK}$ denotes vote-ranked, descriptor-based diversity selection, and $\operatorname{Nbr}(\cdot)$ adds the temporal neighbors of each selected unit.
With long-cluster compression, we additionally select
\begin{equation}
\label{eqn:selected_cluster_evidence}
  \mathcal{E}_{\mathrm{cl}}(q)
  =
  \begin{cases}
    \varnothing, & g(q)=0,\\
    \operatorname{TopK}_{K_{\ell}}
    (\mathcal{L}_T;\,\bm{Q}_q^{P-1},\bm{c}), & g(q)=1,
  \end{cases}
\end{equation}
where cluster centroids are scored against the final shallow query state.
After deduplication and temporal ordering, the gate decides \emph{whether} to retrieve, while token voting, diversity selection, and centroid retrieval decide \emph{what} to retrieve.

\paragraph{Evidence-Based Answering.}
Shallow KVs support query contextualization and retrieval, but not generation.
We gather $\bm{H}_t^0$ for each selected detailed unit and $\overline{\bm{H}}_j^0$ for each selected cluster without recovering its members, compose them with the question, and re-prefill the result through all $L$ language layers:
\begin{equation}
\label{eqn:full_depth_generation}
  \begin{aligned}
    &\widetilde{\bm{H}}^{0}
      =\operatorname{Compose}(
        \{\bm{H}_t^0\}_{t\in\mathcal{E}_{\mathrm{det}}(q)},
        \{\overline{\bm{H}}_j^0\}_{j\in\mathcal{E}_{\mathrm{cl}}(q)},q
      ),\quad
    \widetilde{\bm{H}}^{i+1}
      =F^{i}(\widetilde{\bm{H}}^{i}),\\
    & \textstyle \hat{\bm{y}}
      =\arg\max_{\bm{y}}
      p_{\Theta}\!(
        \bm{y}\mid\widetilde{\bm{H}}^{L}
      ).
  \end{aligned}
\end{equation}
This re-prefill ensures that generation uses consistent full-depth representations.
Only the selected detailed units and cluster representatives receive full-depth computation, so host-to-device transfer is bounded by the evidence budget.
Unselected history remains accessible through the shallow index without requiring full-depth prefill of the entire stream.
All transfer and context-assembly costs are included in the reported end-to-end query latency.




\section{Experiments}
\label{sect:expt}

\subsection{Experimental Setup}
\label{sect:expt:setup}

\paragraph{Datasets and Metrics.}
Following prior streaming-video studies~\citep{Di2025StreamingVQ, Zhang2026HERMESKC}, we evaluate ShallowStream on \textbf{OVO-Bench}~\citep{Li2025OVOBenchHF} and \textbf{StreamingBench}~\citep{Lin2024StreamingBenchAT} using their official protocols and reporting category-wise and aggregate scores.
OVO-Bench separates Real-Time Visual Perception from Backward Tracing, evaluating current-scene understanding and historical retrieval, while StreamingBench covers broader continuous-video scenarios.
We additionally use \textbf{LVBench}~\citep{Wang2024LVBenchAE} to analyze layer-wise retrieval quality and stream-time prefill cost.

\paragraph{Baselines.}
We compare with two baseline groups.
Specialized \textbf{open-sourced online MLLMs} include VideoLLM-online~\citep{Chen2024VideoLLMonlineOV}, Flash-VStream~\citep{Zhang2025FlashVstreamER}, Dispider~\citep{Qian2025DispiderEV}, TimeChat-Online~\citep{Yao2025TimeChatOnline8V}, StreamForest~\citep{Zeng2025StreamForestEO}, and Streamo~\citep{Xia2025StreamingVI}, all designed specifically for causal video processing.
More directly comparable are training-free \textbf{offline-to-online methods}. ReKV~\citep{Di2025StreamingVQ}, StreamKV~\citep{Chen2025StreamKVSV}, and LiveVLM~\citep{Ning2025LiveVLMEO} retrieve historical full-depth or compressed states;  HERMES~\citep{Zhang2026HERMESKC} organizes retained KVs hierarchically; CausalMem~\citep{Song2026TowardsAD} maintains a dynamic fixed-budget memory; and SimpleStream~\citep{Shen2026ASB} uses a strong recent-window baseline. OASIS~\citep{Liang2026OASISOH} and WeaveTime~\citep{Zhang2026WeaveTimeSF} conditionally activate historical retrieval.
Together, these baselines cover recent-only context, KV retention and retrieval, memory compression, fixed-budget memory, and on-demand retrieval.
We also report the original \textbf{LLaVA-OneVision-7B}~\citep{Li2024LLaVAOneVisionEV} and \textbf{Qwen3-VL-8B}~\citep{Bai2025Qwen3VLTR} backbones to contextualize gains within each model family.

\paragraph{Implementation Details.}
We implement ShallowStream in PyTorch using Hugging Face Transformers and FlashAttention-2, with Qwen3-VL-8B-Instruct~\citep{Bai2025Qwen3VLTR} and LLaVA-OneVision-7B~\citep{Li2024LLaVAOneVisionEV} as backbones.
Videos are sampled at 1 FPS for both backbones on both benchmarks. All gate and retrieval settings are fixed before evaluation.
Appendix~\ref{appendix:query_gate_calibration} details gate calibration, and Appendix~\ref{appendix:implementation_settings} provides the complete configuration.

\subsection{Overall Performance and System Efficiency}
\label{sect:expt:overall}

\begin{table}[t]
\centering
\small
\setlength{\tabcolsep}{4.0pt}
\renewcommand{\arraystretch}{1.3}
\caption{\textbf{Main results on OVO-Bench.} Baseline results marked with $^\ddagger$ are from our reruns. Results marked with $^\dagger$ enable long-cluster compression. Dashes indicate unreported entries.}
\label{tab:ovobench_main_results}
\resizebox{0.99\textwidth}{!}{%
\begin{tabular}{ll ccccccc cccc c}
  \toprule
  \multirow[c]{2}{*}[-0.2475em]{\textbf{Model / Method}}
  & \multirow[c]{2}{*}[-0.2475em]{\textbf{\# Frames}}
  & \multicolumn{7}{c}{\textbf{\textit{Real-Time Visual Perception}}}
  & \multicolumn{4}{c}{\textbf{\textit{Backward Tracing}}}
  & \multirow{2}{*}{\textbf{Avg. $\uparrow$}} \\
  \cmidrule(r){3-9}
  \cmidrule(lr){10-13}
  & & \textbf{OCR} & \textbf{ACR} & \textbf{ATR} & \textbf{STU} & \textbf{FPD} & \textbf{OJR} & \textbf{Avg. $\uparrow$}
  & \textbf{EPM} & \textbf{ASI} & \textbf{HLD} & \textbf{Avg. $\uparrow$} & \\

  \midrule
  \rowcolor{black!6} 
  \multicolumn{14}{c}{\textbf{Open-source Online MLLMs}} \\
  \midrule
  VideoLLM-online-8B~\citep{Chen2024VideoLLMonlineOV}
  & 2 fps
  & 8.1 & 23.9 & 12.1 & 14.0 & 45.5 & 21.2 & 20.8
  & 22.2 & 18.8 & 12.2 & 17.7 & 19.3 \\
  Flash-VStream-7B~\citep{Zhang2025FlashVstreamER}
  & 1 fps
  & 25.5 & 32.1 & 29.3 & 33.7 & 29.7 & 28.8 & 29.9
  & 36.4 & 33.8 & 5.9 & 25.4 & 27.6 \\
  Dispider-7B~\citep{Qian2025DispiderEV}
  & 1 fps
  & 57.7 & 49.5 & 62.1 & 44.9 & 61.4 & 51.6 & 54.6
  & 48.5 & 55.4 & 4.3 & 36.1 & 45.3 \\
  TimeChat-Online-7B~\citep{Yao2025TimeChatOnline8V}
  & 1 fps
  & 75.2 & 46.8 & 70.7 & 47.8 & 69.3 & 61.4 & 61.9
  & 55.9 & 59.5 & 9.7 & 41.7 & 51.8 \\
  StreamForest-7B~\citep{Zeng2025StreamForestEO}
  & 1 fps
  & 68.5 & 53.2 & 71.6 & 47.8 & 65.4 & 60.9 & 61.2
  & 58.9 & 64.9 & 32.3 & 52.0 & 56.6 \\
  Streamo-7B~\citep{Xia2025StreamingVI}
  & 1 fps
  & 79.2 & 57.8 & 75.0 &49.4 & 64.4 & 70.1 & 66.0 &54.6 &52.0 &31.7 & 46.1 & 56.1 \\

  \midrule
  \rowcolor{black!6} 
  \multicolumn{14}{c}{\textbf{Offline-to-Online Methods}} \\
  \midrule
  LLaVA-OneVision-7B~\citep{Li2024LLaVAOneVisionEV}
  & 32
  & 66.4&53.2&71.6&51.1&72.3&59.2&62.3&53.9&52.7&18.8&41.8&52.1 \\
  \quad + ReKV~\citep{Di2025StreamingVQ}
  & 0.5 fps
  &52.4&54.1&69.8&43.3&67.3&57.1&57.3&57.6&56.1&18.8&44.2&50.8 \\
  \quad + HERMES$^{\ddagger}$~\citep{Zhang2026HERMESKC}
  & 0.5 fps
  &71.1&61.5&74.1&52.3&72.3&66.3&66.3&62.0&60.8&26.9&\textbf{49.9}&58.1 \\
  \quad + WeaveTime~\citep{Zhang2026WeaveTimeSF}
  & 1 fps
  &72.5&69.7&74.1&53.4&75.2&67.9&68.8&--&--&--&--&-- \\
  \quad + CausalMem~\citep{Song2026TowardsAD}
  & 0.5 fps
  &71.8&61.5&76.7&48.9&76.2&66.8&65.7&--&--&--&--&-- \\
  \quad + SimpleStream$^{\ddagger}$~\citep{Shen2026ASB}
  & 4
  &77.2&71.6&75.9&52.3&77.2&71.7&\underline{71.0}&53.2&52.0&43.6&\underline{49.6}&\underline{60.3} \\
  \rowcolor{tableOnline} 
  \quad + \textbf{ShallowStream (Ours)}
  & 1 fps
  &87.3&73.4&77.6&62.9&74.3&77.2&75.4&50.2&50.7&46.2&49.0&62.2 \\
  \rowcolor{tableOnline} 
  \quad + \textbf{ShallowStream (Ours)}$^{\dagger}$
  & 1 fps
  &87.3&73.4&77.6&63.5&74.3&77.2&\textbf{75.5}&50.5&50.7&46.2&49.1&\textbf{62.3} \\

  \midrule

  Qwen3-VL-8B~\citep{Bai2025Qwen3VLTR}
  & 64
  & 76.5&58.7&75.0&59.0&68.3&59.2&66.1&53.2&66.9&9.7&43.3&54.7 \\
  \quad + ReKV$^{\ddagger}$~\citep{Di2025StreamingVQ}
  & 1 fps
  & 81.9 & 68.8 & 71.6 & 59.0 & 74.3 & 69.0 & 70.8 & 58.6 & 71.6 & 22.6 & 50.9 & 60.8 \\
  \quad + HERMES$^{\ddagger}$~\citep{Zhang2026HERMESKC}
  & 2 fps
  &85.2&64.2&73.3&55.6&70.3&65.8&69.1&50.5&65.5&19.9&45.3&57.2 \\
  \quad + OASIS~\citep{Liang2026OASISOH}
  & --
  &92.0 &80.7 & 81.0 & 67.4 & 67.3 & 79.9 & 78.1 & 62.0 & 60.1 & 47.3 & \underline{57.2} & \underline{67.7} \\
  \quad + SimpleStream$^{\ddagger}$~\citep{Shen2026ASB}
  & 4
  & 92.0 & 81.7 & 81.9 & 69.7 & 75.3 & 81.0 & \underline{80.2} & 53.2 & 60.1 & 44.1 & 52.5 & 66.4 \\
  \rowcolor[HTML]{D9EAD3}  
  \quad + \textbf{ShallowStream (Ours)}
  & 1 fps
  & 92.0 & 83.5 & 82.8 & 71.9 & 75.3 & 79.9 & 80.9 & 52.5 & 72.3 & 49.5 & \textbf{58.1} & \textbf{69.5} \\
  \rowcolor[HTML]{D9EAD3}  
  \quad + \textbf{ShallowStream (Ours)}$^{\dagger}$
  & 1 fps
  & 92.0 & 83.5 & 82.8 & 71.9 & 76.2 & 79.9 & \textbf{81.0} & 51.2 & 71.0 & 50.0 & 57.4 & 69.2 \\
  \bottomrule
\end{tabular}%
}
\end{table}

\begin{table}[t]
\centering
\small
\setlength{\tabcolsep}{4.0pt}
\renewcommand{\arraystretch}{1.3}
\caption{\textbf{Main results on real-time visual understanding subset of StreamingBench.} Baseline results marked with $^\ddagger$ are from our controlled reruns. Results marked with $^\dagger$ enable long-cluster compression. Dashes indicate unreported results.}
\label{tab:streamingbench_main_results}
\begin{tabular}{ll ccccc ccccc c}
\toprule
\multirow[c]{2}{*}[-0.2475em]{\textbf{Model / Method}}
  & \multirow[c]{2}{*}[-0.2475em]{\textbf{\# Frames}}
  & \multicolumn{10}{c}{\textbf{\textit{Real-Time Visual Understanding}}}
  & \multirow{2}{*}{\textbf{Avg. $\uparrow$}} \\
\cmidrule(r){3-12}
  & & \textbf{OP} & \textbf{CR} & \textbf{CS} & \textbf{ATP} & \textbf{EU} & \textbf{TR} & \textbf{PR} & \textbf{SU} & \textbf{ACP} & \textbf{CT} & \\
\midrule
\rowcolor{black!6} 
\multicolumn{13}{c}{\textbf{Open-source Online MLLMs}} \\
\midrule
Flash-VStream-7B~\citep{Zhang2025FlashVstreamER}
  & -- & 25.9 & 43.6 & 24.9 & 23.9 & 27.3 & 13.1 & 18.5 & 25.2 & 23.9 & 48.7 & 23.2 \\
VideoLLM-online-8B~\citep{Chen2024VideoLLMonlineOV}
  & 2 fps & 39.1 & 40.1 & 34.5 & 31.1 & 46.0 & 32.4 & 31.5 & 34.2 & 42.5 & 27.9 & 36.0 \\
Dispider-7B~\citep{Qian2025DispiderEV}
  & 1 fps & 74.9 & 75.5 & 74.1 & 73.1 & 74.4 & 59.9 & 76.1 & 62.9 & 62.2 & 45.8 & 67.6 \\
TimeChat-Online-7B~\citep{Yao2025TimeChatOnline8V}
  & 1 fps & 80.2 & 82.0 & 79.5 & 83.3 & 76.1 & 78.5 & 78.7 & 64.6 & 69.6 & 58.0 & 75.4 \\
StreamForest-7B~\citep{Zeng2025StreamForestEO}
  & 1 fps & 83.1 & 82.8 & 82.7 & 84.3 & 77.5 & 78.2 & 76.9 & 69.1 & 75.6 & 54.4 & 77.3 \\

\midrule
\rowcolor{black!6} 
\multicolumn{13}{c}{\textbf{Offline-to-Online Methods}} \\
\midrule
LLaVA-OneVision-7B~\citep{Li2024LLaVAOneVisionEV}
  & 32 & 77.7&76.6&77.6&81.9&71.7&71.7&66.7&65.5&65.7&44.0&71.0 \\

\quad + ReKV~\citep{Di2025StreamingVQ,Zhang2026HERMESKC}
  & 0.5 fps
  & 74.4&78.9&78.6&77.1&68.3&67.9&67.6&62.6&64.3&44.6&69.2 \\

\quad + HERMES$^{\ddagger}$~\citep{Zhang2026HERMESKC}
  & 0.5 fps
  & 78.2&79.7&86.8&81.5&69.2&73.2&75.0&64.2&68.6&45.6&73.2 \\

\quad + StreamKV~\citep{Chen2025StreamKVSV}
  & 1 fps
  & 74.7&78.1&87.7&79.4&70.8&67.6&70.4&64.6&64.0&45.1&71.0 \\

\quad + WeaveTime~\citep{Zhang2026WeaveTimeSF}
  & 1 fps
  & 71.5&81.3&86.8&78.6&75.2&73.2&72.2&69.1&68.8&44.7&72.1 \\

\quad + SimpleStream$^{\ddagger}$~\citep{Shen2026ASB}
  & 4
  & 80.4&71.9&85.2&85.5&76.1&73.8&69.4&65.5&71.1&37.3&73.5 \\

\quad + LiveVLM~\citep{Ning2025LiveVLMEO}
  & 0.5 fps
  & 79.8&79.7&84.9&80.7&67.1&70.1&74.1&66.3&68.0&42.5&71.3 \\

\quad + CausalMem~\citep{Song2026TowardsAD}
  & 0.5 fps
  & 82.6&79.7&83.2&83.2&69.6&78.4&75.0&67.3&70.9&39.4&\underline{74.3} \\


\rowcolor{tableOnline} 
\quad + \textbf{ShallowStream (Ours)}
  & 1 fps
  & 82.8&67.2&83.9&90.8&71.7&81.9&64.8&70.7&74.8&35.2&75.5 \\

\rowcolor{tableOnline} 
\quad + \textbf{ShallowStream (Ours)}$^{\dagger}$
  & 1 fps
  & 82.8&67.2&83.9&90.8&72.3&81.9&64.8&70.7&74.8&35.2&\textbf{75.6} \\

\midrule
Qwen3-VL-8B~\citep{Bai2025Qwen3VLTR}
  & 64 & 80.4&75.0&83.0&83.5&74.2&84.4&77.8&63.8&69.7&57.0&75.9 \\

\quad + SimpleStream$^{\ddagger}$~\citep{Shen2026ASB}
  & 4
  & 79.3&72.7&89.9&84.8&73.6&79.1&83.3&71.1&76.8&39.4&76.5 \\

\quad + HERMES$^{\ddagger}$~\citep{Zhang2026HERMESKC}
  & 2 fps
  & 79.0&77.3&81.4&84.2&73.6&77.9&87.0&70.7&72.5&61.1&\underline{76.6} \\


\rowcolor[HTML]{D9EAD3}  
\quad + \textbf{ShallowStream (Ours)}
  & 1 fps
  & 79.3&71.9&89.9&85.8&77.4&79.1&88.9&71.5&77.3&52.9&\textbf{78.2} \\

\rowcolor[HTML]{D9EAD3}  
\quad + \textbf{ShallowStream (Ours)}$^{\dagger}$
  & 1 fps
  & 79.3&71.9&90.2&85.5&77.4&79.1&88.9&71.5&77.3&49.7&78.0 \\
\bottomrule
\end{tabular}
\end{table}

\paragraph{Overall Streaming Performance.}
Tables~\ref{tab:ovobench_main_results} and~\ref{tab:streamingbench_main_results} evaluate two complementary capabilities: retrieving evidence from the observed history and understanding an evolving visual scene.
Without task-specific training, ShallowStream achieves the best overall results for both backbones: \textbf{69.5}/\textbf{78.2} on OVO-Bench/StreamingBench with Qwen3-VL-8B and \textbf{62.2}/\textbf{75.5} with LLaVA-OneVision-7B.
Across the four dataset-backbone settings, it outperforms the second-best same-backbone method by \textbf{1.7 points} on average, showing that shallow stream-time indexing preserves evidence for both retrospective and real-time questions across different MLLM backbones.

\paragraph{Continuous-Stream Compute Efficiency.}
Efficiency is measured on one NVIDIA RTX 5090 with Qwen3-VL-8B over five long OVO-Bench Backward videos at 1 FPS.
Timing excludes frame decoding and fixes generation to 16 visible tokens.
We separately report per-frame prefill and per-query computation, including ShallowStream's archive updates, selected-state transfer, context assembly, and generation.
Memory results report peak GPU allocation, excluding CPU capacity.

ShallowStream achieves performance on par with current state-of-the-art streaming methods at substantially lower continuous-stream cost.
As shown in Figure~\ref{fig:latency-performance-teaser}, it reduces per-frame prefill latency by up to \textbf{52.1$\times$} and 20-s end-to-end latency by up to \textbf{15.3$\times$}.
These gains reflect the two central design choices in Section~\ref{sect:method}: processing each incoming unit only through the shallow layers and reserving full-depth computation for the evidence selected at query time.
Appendix~\ref{appendix:pruning_boundary_sensitivity} analyzes the trade-off between retrieval quality and stream-processing cost as the pruning boundary varies.

\begin{figure}[t]
  \centering
  \includegraphics[width=0.99\linewidth]{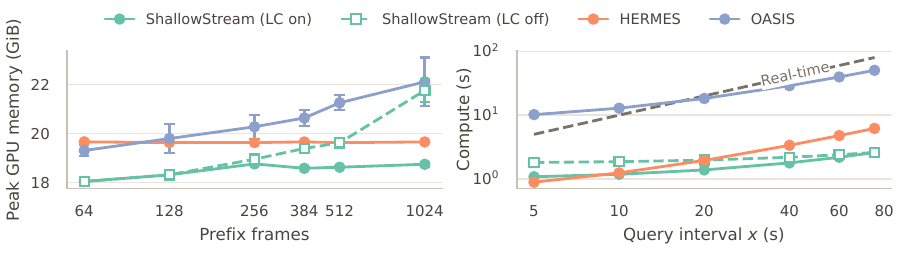}
  \vspace{-0.5em}
  \caption{\textbf{Memory scaling and real-time compute demand.}
  \emph{Left:} Peak GPU allocation from 64 to 1,024 frames, averaged over five videos; both ShallowStream variants use the calibrated Gate and token-vote retriever.
  \emph{Right:} Total compute for a query interval $x$, obtained by combining mean query computation with stream-time prefill accumulated over $x$ seconds at 1 FPS.
  The dashed boundary $y=x$ separates configurations that keep pace with the stream (below) from those whose compute exceeds the available interval (above).}
  \label{fig:peak-memory-growth}
\end{figure}

\paragraph{Memory Scaling under Long-History Compression.}
Long-history compression (Section~\ref{sect:method:stream_encoding}) bounds GPU-memory growth by replacing older shallow states with fixed-size cluster representatives.
The left panel of Figure~\ref{fig:peak-memory-growth} compares ShallowStream with compression enabled (\textbf{LC-on}) and disabled (\textbf{LC-off}); LC-off retains the complete layer-$0$--$4$ shallow K/V history.
From 64 to 1,024 frames, LC-on keeps peak GPU allocation nearly constant at approximately 18 GiB, whereas LC-off grows to 21.76 GiB.
At long prefixes, LC-on also uses less GPU memory than HERMES and OASIS.

\paragraph{Real-Time Processing Headroom.}
The right panel of Figure~\ref{fig:peak-memory-growth} evaluates total compute demand over different query intervals on five videos containing 588--1,198 sampled frames, without imposing a frame cap.
Both LC-on and LC-off remain well below the real-time boundary.
At an 80-s query interval, each requires approximately $2.6$ s of compute, compared with $6.2$ s for HERMES and $50.4$ s for OASIS.
Thus, ShallowStream retains substantial processing headroom whether it compresses old history or preserves the complete shallow index.
Appendix~\ref{appendix:query_history_scaling} further examines how LC-off scales with the retained history.

\subsection{Routing and Retrieval Ablations}
\label{sect:expt:ablation}

\begin{figure}[t]
  \centering
  \begin{minipage}[t]{0.5\textwidth}
    \centering
    \includegraphics[width=0.99\linewidth]{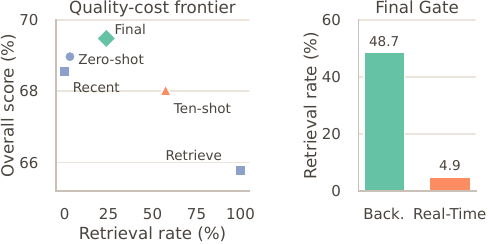}
    \caption{\textbf{Retrieval-gate ablation on OVO-Bench.} \emph{Left}: Accuracy vs. retrieval rate with fixed branch outputs, isolating routing effects; \emph{Right}: Calibrated Gate retrieval rates for Backward and Real-Time questions.}
    \label{fig:gate-causal-ablation}
    \vspace{-0.5em}
  \end{minipage}
  \hfill
  \begin{minipage}[t]{0.46\textwidth}
    \centering
    \includegraphics[width=0.99\linewidth]{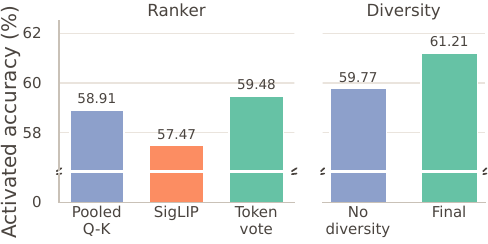}
    \caption{\textbf{Evidence-retrieval ablation on Gate-selected OVO-Bench queries.} \emph{Left:} Rankers without diversity or temporal expansion; \emph{Right:} Max-min diversity with token voting and preceding-unit expansion fixed.}
    \label{fig:evidence-ranker-ablation}
    \vspace{-0.5em}
  \end{minipage}
\end{figure}

\paragraph{The Retrieval Gate Identifies Historical Need.}
Always using recent context can omit necessary historical evidence, whereas indiscriminate retrieval can introduce distracting visual content.
Figure~\ref{fig:gate-causal-ablation} shows that the Query-Logit Gate from Section~\ref{sect:method:query_answering} balances these cases by activating retrieval for 48.7\% of Backward questions but only 4.9\% of Real-Time questions.
Under controlled branch outputs, this selective routing improves overall accuracy over both always-recent and always-retrieve policies.
The result confirms that the Gate responds to the question's historical requirements rather than applying retrieval uniformly.

\paragraph{Token Voting and Diversity Improve Evidence Selection.}
Figure~\ref{fig:evidence-ranker-ablation} isolates the evidence-retrieval components introduced in Section \ref{sect:method:query_answering}.
Without diversity selection or temporal expansion, cross-layer token voting achieves 59.48\%, outperforming pooled shallow Q-K (58.91\%) and the independent SigLIP encoder (57.47\%).
This improvement demonstrates the value of query-contextualized, token-level attention for locating historical evidence.
With token voting and preceding-unit expansion fixed, max--min diversity further raises accuracy from 59.77\% to 61.21\% by allocating the limited evidence budget across complementary moments.

\begin{figure}[!t]
  \centering
  \includegraphics[width=0.99\linewidth]{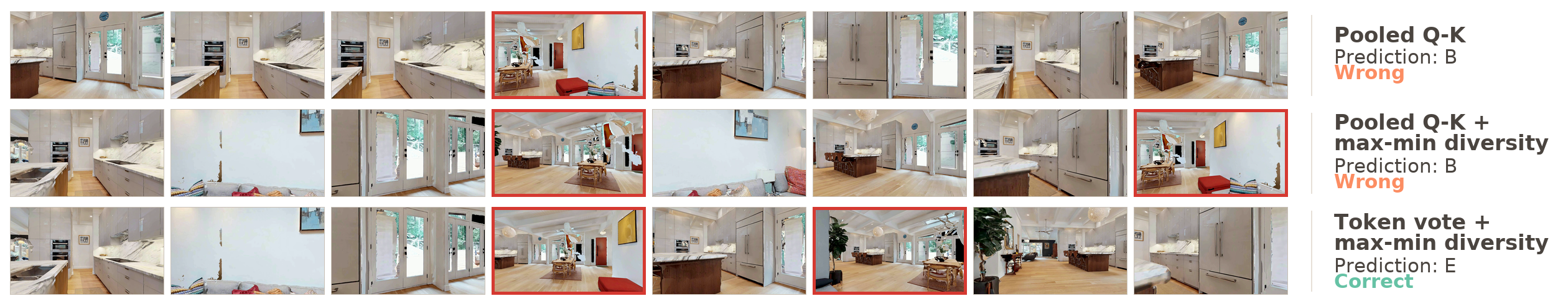}
  \caption{\textbf{Qualitative evidence retrieval on OVO-Bench EPM.}
  For the question ``Where is the red and white checkered rug?'' (ground-truth answer: E), the rows show the eight units selected by pooled shallow Q-K, pooled shallow Q-K with max-min diversity, and the final token-vote retriever with max-min diversity. Red outlines indicate frames containing the rug beneath the dining table. The final retriever selects complementary views and produces the correct answer.}
  \label{fig:retrieval-case-study}
\end{figure}

\subsection{Qualitative Evidence Retrieval}
\label{sect:expt:case}

Figure~\ref{fig:retrieval-case-study}  illustrates how the final retriever combines the two evidence-selection mechanisms in Section~\ref{sect:method}.
Pooled shallow Q-K repeatedly selects similar kitchen views and predicts B. Max-min diversity broadens its temporal coverage, but the selections remain constrained by the pooled relevance signal and yield the same answer.
Cross-layer token voting instead identifies query-specific evidence, while diversity selection distributes the retrieval budget across complementary views of the room, enabling the model to answer E correctly.

\section{Conclusions}
\label{sect:conclusion}

We introduced \textbf{ShallowStream}, which treats model depth as a resource to allocate selectively in streaming video understanding.
Rather than processing every incoming frame through the full MLLM, it builds a lightweight shallow index and applies full-depth computation only to evidence selected through query-logit routing and diversity-aware token voting.
Across two MLLM backbones, ShallowStream matches state-of-the-art performance while substantially reducing computation, latency, and GPU-memory growth.
These results establish shallow indexing with selective deep processing as a practical design principle for efficient, long-running streaming MLLMs.

\subsection*{AI Use Statement}

Generative AI tools were used to generate the benchmark-independent synthetic questions and routing labels used for Gate calibration, and to polish the manuscript. The authors reviewed the generated calibration data for consistency with the retrospective-versus-recent routing definition and reviewed all AI-assisted text. The authors take full responsibility for the final content of this work.

\subsection*{Ethics Statement}

This work develops a computational method for streaming video understanding and evaluates it using existing pretrained models and public benchmarks. It does not involve human-subject research, the collection of personal or sensitive data, or deployment in safety-critical settings. We identify no specific ethical concerns arising from the proposed method.

\subsection*{Reproducibility Statement}

Implementation details, model and benchmark configurations, calibration procedures, and evaluation protocols are documented in the main paper and appendix. The source code and experiment configurations are available at \url{https://anonymous.4open.science/r/ShallowStream/}. Together, these materials specify the data processing, inference settings, and measurements needed to reproduce the reported results.


\bibliographystyle{plainnat}
\bibliography{reference}

@inproceedings{Di2025StreamingVQ,
  title = {Streaming Video Question-Answering with In-context Video KV-Cache Retrieval},
  author = {Di, Shangzhe and Yu, Zhelun and Zhang, Guanghao and Li, Haoyuan and Zhong, Tao and Cheng, Hao and Li, Bolin and He, Wanggui and Shu, Fangxun and Jiang, Hao},
  booktitle = {International Conference on Learning Representations},
  year = {2025},
  volume = {2025},
  pages = {42115--42127},
  url = {https://api.semanticscholar.org/CorpusID:276741475}
}

@inproceedings{Chen2025StreamKVSV,
  title = {StreamKV: Streaming Video Question-Answering with Segment-based KV Cache Retrieval and Compression},
  author = {Chen, Yilong and Bai, Xiang and Wang, Zhibin and Bai, Chengyu and Dai, Yuhan and Lu, Ming},
  booktitle = {Proceedings of the AAAI Conference on Artificial Intelligence},
  year = {2026},
  volume = {40},
  number = {4},
  pages = {3120--3128},
  doi = {10.1609/aaai.v40i4.37305},
  url = {https://api.semanticscholar.org/CorpusID:282913012}
}

@inproceedings{Liang2026OASISOH,
  title = {OASIS: On-Demand Hierarchical Event Memory for Streaming Video Reasoning},
  author = {Liang, Zhijia and Li, Jiaming and Chen, Weikai and Zhang, Yanhao and Lu, Haonan and Li, Guanbin},
  booktitle = {Proceedings of the IEEE/CVF Conference on Computer Vision and Pattern Recognition (CVPR)},
  year = {2026},
  pages = {2821--2831},
  month = {June},
  url = {https://api.semanticscholar.org/CorpusID:287635604}
}

@inproceedings{Chen2023InternVS,
  title = {InternVL: Scaling up Vision Foundation Models and Aligning for Generic Visual-Linguistic Tasks},
  author = {Chen, Zhe and Wu, Jiannan and Wang, Wenhai and Su, Weijie and Chen, Guo and Xing, Sen and Zhong, Muyan and Zhang, Qinglong and Zhu, Xizhou and Lu, Lewei and Li, Bin and Luo, Ping and Lu, Tong and Qiao, Yu and Dai, Jifeng},
  booktitle = {Proceedings of the IEEE/CVF Conference on Computer Vision and Pattern Recognition (CVPR)},
  year = {2024},
  pages = {24185--24198},
  month = {June},
  url = {https://api.semanticscholar.org/CorpusID:266521410}
}

@article{Bai2025Qwen25VLTR,
  title = {Qwen2.5-VL Technical Report},
  author = {Bai, Shuai and Chen, Keqin and Liu, Xuejing and Wang, Jialin and Ge, Wenbin and Song, Sibo and Dang, Kai and Wang, Peng and Wang, Shijie and Tang, Jun and Zhong, Humen and Zhu, Yuanzhi and Yang, Mingkun and Li, Zhaohai and Wan, Jianqiang and Wang, Pengfei and Ding, Wei and Fu, Zheren and Xu, Yiheng and Ye, Jiabo and Zhang, Xi and Xie, Tianbao and Cheng, Zesen and Zhang, Hang and Yang, Zhibo and Xu, Haiyang and Lin, Junyang},
  journal = {ArXiv},
  year = {2025},
  volume = {abs/2502.13923},
  url = {https://api.semanticscholar.org/CorpusID:276449796}
}

@article{Bai2025Qwen3VLTR,
  title = {Qwen3-VL Technical Report},
  author = {Bai, Shuai and Cai, Yuxuan and Chen, Ruizhe and Chen, Keqin and Chen, Xionghui and Cheng, Zesen and Deng, Lianghao and Ding, Wei and Gao, Chang and Ge, Chunjiang and Ge, Wenbin and Guo, Zhifang and Huang, Qidong and Huang, Jie and Huang, Fei and Hui, Binyuan and Jiang, Shutong and Li, Zhaohai and Li, Mingsheng and Li, Mei and Li, Kaixin and Lin, Zicheng and Lin, Junyang and Liu, Xuejing and Liu, Jiawei and Liu, Chenglong and Liu, Yang and Liu, Dayiheng and Liu, Shixuan and Lu, Dunjie and Luo, Ruilin and Lv, Chenxu and Men, Rui and Meng, Lingchen and Ren, Xuancheng and Ren, Xingzhang and Song, Sibo and Sun, Yuchong and Tang, Jun and Tu, Jianhong and Wan, Jianqiang and Wang, Peng and Wang, Pengfei and Wang, Qiuyue and Wang, Yuxuan and Xie, Tianbao and Xu, Yiheng and Xu, Haiyang and Xu, Jin and Yang, Zhibo and Yang, Mingkun and Yang, Jianxin and Yang, An and Yu, Bowen and Zhang, Fei and Zhang, Hang and Zhang, Xi and Zheng, Bo and Zhong, Humen and Zhou, Jingren and Zhou, Fan and Zhou, Jing and Zhu, Yuanzhi and Zhu, Ke},
  journal = {ArXiv},
  year = {2025},
  volume = {abs/2511.21631},
  url = {https://api.semanticscholar.org/CorpusID:283262018}
}

@inproceedings{Chen2024VideoLLMonlineOV,
  title = {VideoLLM-online: Online Video Large Language Model for Streaming Video},
  author = {Chen, Joya and Lv, Zhaoyang and Wu, Shiwei and Lin, Kevin Qinghong and Song, Chenan and Gao, Difei and Liu, Jia-Wei and Gao, Ziteng and Mao, Dongxing and Shou, Mike Zheng},
  booktitle = {Proceedings of the IEEE/CVF Conference on Computer Vision and Pattern Recognition (CVPR)},
  year = {2024},
  pages = {18407--18418},
  month = {June},
  url = {https://api.semanticscholar.org/CorpusID:270560262}
}

@inproceedings{Qian2024StreamingLV,
  title = {Streaming Long Video Understanding with Large Language Models},
  author = {Qian, Rui and Dong, Xiaoyi and Zhang, Pan and Zang, Yuhang and Ding, Shuangrui and Lin, Dahua and Wang, Jiaqi},
  booktitle = {Advances in Neural Information Processing Systems},
  year = {2024},
  volume = {37},
  pages = {119336--119360},
  publisher = {Curran Associates, Inc.},
  doi = {10.52202/079017-3792},
  url = {https://api.semanticscholar.org/CorpusID:270063250}
}

@inproceedings{Li2025OVOBenchHF,
  title = {OVO-Bench: How Far is Your Video-LLMs from Real-World Online Video Understanding?},
  author = {Niu, Junbo and Li, Yifei and Miao, Ziyang and Ge, Chunjiang and Zhou, Yuanhang and He, Qihao and Dong, Xiaoyi and Duan, Haodong and Ding, Shuangrui and Qian, Rui and Zhang, Pan and Zang, Yuhang and Cao, Yuhang and He, Conghui and Wang, Jiaqi},
  booktitle = {Proceedings of the IEEE/CVF Conference on Computer Vision and Pattern Recognition (CVPR)},
  year = {2025},
  pages = {18902--18913},
  month = {June},
  url = {https://api.semanticscholar.org/CorpusID:275458756}
}

@inproceedings{Lin2024StreamingBenchAT,
  title = {StreamingBench: Assessing the Gap for MLLMs to Achieve Streaming Video Understanding},
  author = {Junming Lin and Zheng Fang and Chi Chen and Haoxuan Cheng and Zihao Wan and Fuwen Luo and Ziyue Wang and Peng Li and Yang Liu and Maosong Sun},
  booktitle = {ICASSP 2026 - 2026 IEEE International Conference on Acoustics, Speech and Signal Processing (ICASSP)},
  year = {2026},
  pages = {12147--12151},
  url = {https://api.semanticscholar.org/CorpusID:273850338}
}

@inproceedings{Chen2025StreamingTOMST,
  title = {StreamingTOM: Streaming Token Compression for Efficient Video Understanding},
  author = {Chen, Xueyi and Tao, Keda and Shao, Kele and Wang, Huan},
  booktitle = {Proceedings of the IEEE/CVF Conference on Computer Vision and Pattern Recognition (CVPR)},
  year = {2026},
  pages = {24675--24685},
  month = {June},
  url = {https://api.semanticscholar.org/CorpusID:282246334}
}

@article{Shen2026ASB,
  title = {A Simple Baseline for Streaming Video Understanding},
  author = {Shen, Yujiao and Tian, Shulin and Yang, Jingkang and Liu, Ziwei},
  journal = {ArXiv},
  year = {2026},
  volume = {abs/2604.02317},
  url = {https://api.semanticscholar.org/CorpusID:287073889}
}

@inproceedings{Zhang2026WeaveTimeSF,
  title = {WeaveTime: Streaming from Earlier Frames into Emergent Memory in VideoLLMs},
  author = {Zhang, Yulin and Shi, Cheng and Yang, Sibei},
  booktitle = {Proceedings of the IEEE/CVF Conference on Computer Vision and Pattern Recognition (CVPR)},
  year = {2026},
  pages = {16920--16932},
  month = {June},
  url = {https://api.semanticscholar.org/CorpusID:286011636}
}

@inproceedings{Yang2025StreamMemQK,
  title = {StreamMem: Query-Agnostic KV Cache Memory for Streaming Video Understanding},
  author = {Yanlai Yang and Zhuokai Zhao and Satya Narayan Shukla and Aashu Singh and Shlok Kumar Mishra and Lizhu Zhang and Mengye Ren},
  booktitle = {CVPR Workshop on Video LLMs},
  year = {2026},
  url = {https://api.semanticscholar.org/CorpusID:280700196}
}

@article{Song2026TowardsAD,
  title = {Towards a Dynamic and Fixed-budget Memory Bank for Efficient Streaming Video Understanding},
  author = {Song, Baiyang and Lin, Yuli and Wu, Qiong and Chen, Tao and Peng, Jun and Chen, Xiao and Zhou, Yiyi and Ji, Rongrong},
  journal = {ArXiv},
  year = {2026},
  volume = {abs/2606.25658},
  url = {https://api.semanticscholar.org/CorpusID:289630861}
}

@inproceedings{Yao2025TimeChatOnline8V,
  title = {TimeChat-Online: 80\% Visual Tokens are Naturally Redundant in Streaming Videos},
  author = {Yao, Linli and Li, Yicheng and Wei, Yuancheng and Li, Lei and Ren, Shuhuai and Liu, Yuanxin and Ouyang, Kun and Wang, Lean and Li, Shicheng and Li, Sida and Kong, Lingpeng and Liu, Qi and Zhang, Yuanxing and Sun, Xu},
  booktitle = {Proceedings of the 33rd ACM International Conference on Multimedia},
  year = {2025},
  pages = {10807--10816},
  doi = {10.1145/3746027.3754839},
  url = {https://api.semanticscholar.org/CorpusID:278033234}
}

@inproceedings{Kim2025InfiniPotVMK,
  title = {InfiniPot-V: Memory-Constrained KV Cache Compression for Streaming Video Understanding},
  author = {Kim, Minsoo and Shim, Kyuhong and Choi, Jungwook and Chang, Simyung},
  booktitle = {Advances in Neural Information Processing Systems},
  year = {2025},
  volume = {38},
  pages = {138983--139013},
  publisher = {Curran Associates, Inc.},
  doi = {10.52202/085713-4639},
  url = {https://api.semanticscholar.org/CorpusID:279465408}
}

@inproceedings{Ning2025LiveVLMEO,
  title = {LiveVLM: Efficient Online Video Understanding via Streaming-Oriented KV Cache and Retrieval},
  author = {Zhenyu Ning and Guangda Liu and Qihao Jin and Chengwei Li and Wenchao Ding and Minyi Guo and Jieru Zhao},
  booktitle = {Proceedings of the 63rd ACM/IEEE Design Automation Conference},
  year = {2026},
  doi = {10.1145/3770743.3804012},
  url = {https://api.semanticscholar.org/CorpusID:278783116}
}

@inproceedings{Xiao2026MuKVMK,
  title = {MuKV: Multi-Grained KV Cache Compression for Long Streaming Video Question-Answering},
  author = {Xiao, Junbin and Chen, Jiajun and Sun, Tianxiang and Yang, Xun and Yao, Angela},
  booktitle = {Proceedings of the IEEE/CVF Conference on Computer Vision and Pattern Recognition (CVPR)},
  year = {2026},
  pages = {11381--11391},
  month = {June},
  url = {https://api.semanticscholar.org/CorpusID:288650113}
}

@inproceedings{Li2024SCBenchAK,
  title = {SCBench: A KV Cache-Centric Analysis of Long-Context Methods},
  author = {Li, Yucheng and Jiang, Huiqiang and Wu, Qianhui and Luo, Xufang and Ahn, Surin and Zhang, Chengruidong and Abdi, Amir and Li, Dongsheng and Gao, Jianfeng and Yang, Yuqing and Qiu, Lili},
  booktitle = {International Conference on Learning Representations},
  year = {2025},
  volume = {2025},
  pages = {66063--66093},
  url = {https://api.semanticscholar.org/CorpusID:274763141}
}

@inproceedings{Kim2025VRexRS,
  title = {V-Rex: Real-Time Streaming Video LLM Acceleration via Dynamic KV Cache Retrieval},
  author = {Kim, Donghyuk and Yang, Sejeong and Shin, Wonjin and Kim, Joo-Young},
  booktitle = {2026 IEEE International Symposium on High Performance Computer Architecture (HPCA)},
  year = {2026},
  pages = {1--14},
  doi = {10.1109/hpca68181.2026.11408603},
  url = {https://api.semanticscholar.org/CorpusID:283895999}
}

@inproceedings{Xin2020EarlyEB,
  title = {Early Exiting {BERT} for Efficient Document Ranking},
  author = {Xin, Ji and Nogueira, Rodrigo and Yu, Yaoliang and Lin, Jimmy},
  booktitle = {Proceedings of SustaiNLP: Workshop on Simple and Efficient Natural Language Processing},
  year = {2020},
  pages = {83--88},
  month = {November},
  publisher = {Association for Computational Linguistics},
  doi = {10.18653/v1/2020.sustainlp-1.11},
  url = {https://api.semanticscholar.org/CorpusID:226283755}
}

@inproceedings{Meng2022LocatingAE,
  title = {Locating and Editing Factual Associations in GPT},
  author = {Meng, Kevin and Bau, David and Andonian, Alex and Belinkov, Yonatan},
  booktitle = {Advances in Neural Information Processing Systems},
  year = {2022},
  volume = {35},
  pages = {17359--17372},
  publisher = {Curran Associates, Inc.},
  doi = {10.52202/068431-1262},
  url = {https://api.semanticscholar.org/CorpusID:255825985}
}

@inproceedings{Wang2024LVBenchAE,
  title = {LVBench: An Extreme Long Video Understanding Benchmark},
  author = {Wang, Weihan and He, Zehai and Hong, Wenyi and Cheng, Yean and Zhang, Xiaohan and Qi, Ji and Ding, Ming and Gu, Xiaotao and Huang, Shiyu and Xu, Bin and Dong, Yuxiao and Tang, Jie},
  booktitle = {Proceedings of the IEEE/CVF International Conference on Computer Vision (ICCV)},
  year = {2025},
  pages = {22958--22967},
  month = {October},
  url = {https://api.semanticscholar.org/CorpusID:270391637}
}

@article{Li2024LLaVAOneVisionEV,
  title = {LLaVA-OneVision: Easy Visual Task Transfer},
  author = {Bo Li and Yuanhan Zhang and Dong Guo and Renrui Zhang and Feng Li and Hao Zhang and Kaichen Zhang and Peiyuan Zhang and Yanwei Li and Ziwei Liu and Chunyuan Li},
  journal = {Transactions on Machine Learning Research},
  year = {2025},
  url = {https://api.semanticscholar.org/CorpusID:271719914}
}

@inproceedings{Wang2024VideoLLaMBLS,
  title = {VideoLLaMB: Long Streaming Video Understanding with Recurrent Memory Bridges},
  author = {Wang, Yuxuan and Song, Yiqi and Xie, Cihang and Liu, Yang and Zheng, Zilong},
  booktitle = {Proceedings of the IEEE/CVF International Conference on Computer Vision (ICCV)},
  year = {2025},
  pages = {24170--24181},
  month = {October},
  url = {https://api.semanticscholar.org/CorpusID:280422760}
}

@inproceedings{Yang2025SVBenchAB,
  title = {SVBench: A Benchmark with Temporal Multi-Turn Dialogues for Streaming Video Understanding},
  author = {Yang, Zhenyu and Hu, Yuhang and Du, Zemin and Xue, Dizhan and Qian, Shengsheng and Wu, Jiahong and Yang, Fan and Dong, Weiming and Xu, Changsheng},
  booktitle = {International Conference on Learning Representations},
  year = {2025},
  volume = {2025},
  pages = {42522--42556},
  url = {https://api.semanticscholar.org/CorpusID:276408575}
}

@inproceedings{Hao2025OmniKVDC,
  title = {OmniKV: Dynamic Context Selection for Efficient Long-Context LLMs},
  author = {Hao, Jitai and Zhu, Yuke and Wang, Tian and Yu, Jun and Xin, Xin and Zheng, Bo and Ren, Zhaochun and Guo, Sheng},
  booktitle = {International Conference on Learning Representations},
  year = {2025},
  volume = {2025},
  pages = {87443--87464},
  url = {https://api.semanticscholar.org/CorpusID:278601790}
}

@inproceedings{Zhang2026HERMESKC,
  title = {{HERMES}: {KV} Cache as Hierarchical Memory for Efficient Streaming Video Understanding},
  author = {Zhang, Haowei and Yang, Shudong and Fu, Jinlan and Ng, See-Kiong and Qiu, Xipeng},
  booktitle = {Proceedings of the 64th Annual Meeting of the {A}ssociation for {C}omputational {L}inguistics (Volume 1: Long Papers)},
  year = {2026},
  pages = {8411--8430},
  month = {July},
  publisher = {Association for Computational Linguistics},
  doi = {10.18653/v1/2026.acl-long.381},
  url = {https://api.semanticscholar.org/CorpusID:284917012}
}

@inproceedings{Lu2026VistaSO,
  title = {Vista: Scene-Aware Optimization for Streaming Video Question Answering under Post-Hoc Queries},
  author = {Haocheng Lu and Nan Zhang and Wei Tao and Xiaoyang Qu and Guokuan Li and Jiguang Wan and Jianzong Wang},
  booktitle = {Proceedings of the AAAI Conference on Artificial Intelligence},
  year = {2026},
  volume = {40},
  number = {9},
  pages = {7539--7547},
  doi = {10.1609/aaai.v40i9.37694},
  url = {https://api.semanticscholar.org/CorpusID:285454172}
}

@inproceedings{Zhang2025FlashVstreamER,
  title = {Flash-VStream: Efficient Real-Time Understanding for Long Video Streams},
  author = {Zhang, Haoji and Wang, Yiqin and Tang, Yansong and Liu, Yong and Feng, Jiashi and Jin, Xiaojie},
  booktitle = {Proceedings of the IEEE/CVF International Conference on Computer Vision (ICCV)},
  year = {2025},
  pages = {21059--21069},
  month = {October},
  url = {https://api.semanticscholar.org/CorpusID:280011608}
}

@inproceedings{Qian2025DispiderEV,
  title = {Dispider: Enabling Video LLMs with Active Real-Time Interaction via Disentangled Perception, Decision, and Reaction},
  author = {Qian, Rui and Ding, Shuangrui and Dong, Xiaoyi and Zhang, Pan and Zang, Yuhang and Cao, Yuhang and Lin, Dahua and Wang, Jiaqi},
  booktitle = {Proceedings of the IEEE/CVF Conference on Computer Vision and Pattern Recognition (CVPR)},
  year = {2025},
  pages = {24045--24055},
  month = {June},
  url = {https://api.semanticscholar.org/CorpusID:275336693}
}

@inproceedings{Zeng2025StreamForestEO,
  title = {StreamForest: Efficient Online Video Understanding with Persistent Event Memory},
  author = {Zeng, Xiangyu and Qiu, Kefan and Zhang, Qingyu and Li, Xinhao and Wang, Jing and Li, Jiaxin and Yan, Ziang and Tian, Kun and Tian, Meng and Zhao, Xinhai and Wang, Yi and Wang, Limin},
  booktitle = {Advances in Neural Information Processing Systems},
  year = {2025},
  volume = {38},
  pages = {75804--75835},
  publisher = {Curran Associates, Inc.},
  doi = {10.52202/085713-2547},
  url = {https://api.semanticscholar.org/CorpusID:281674907}
}

@inproceedings{Xia2025StreamingVI,
  title = {Streaming Video Instruction Tuning},
  author = {Xia, Jiaer and Chen, Peixian and Zhang, Mengdan and Sun, Xing and Zhou, Kaiyang},
  booktitle = {Proceedings of the IEEE/CVF Conference on Computer Vision and Pattern Recognition (CVPR)},
  year = {2026},
  pages = {31219--31229},
  month = {June},
  url = {https://api.semanticscholar.org/CorpusID:284153645}
}

@inproceedings{Wang2025StreamBridgeTY,
  title = {StreamBridge: Turning Your Offline Video Large Language Model into a Proactive Streaming Assistant},
  author = {Wang, Haibo and Feng, Bo and Lai, Zhengfeng and Xu, Mingze and Li, Shiyu and Ge, Weifeng and Dehghan, Afshin and Cao, Meng and Huang, Ping},
  booktitle = {Advances in Neural Information Processing Systems},
  year = {2025},
  volume = {38},
  pages = {132332--132359},
  publisher = {Curran Associates, Inc.},
  doi = {10.52202/085713-4406},
  url = {https://api.semanticscholar.org/CorpusID:278394112}
}

@inproceedings{Wen2026EventMemAgentHE,
  title = {EventMemAgent: Hierarchical Event-Centric Memory for Online Video Understanding with Adaptive Tool Use},
  author = {Siwei Wen and Zhangcheng Wang and Xingjian Zhang and Lei Huang and Wenjun Wu},
  booktitle = {European Conference on Computer Vision (ECCV)},
  year = {2026},
  note = {Accepted for publication},
  url = {https://api.semanticscholar.org/CorpusID:285659749}
}

@inproceedings{Yan2025LearningSV,
  title = {Learning Streaming Video Representation via Multitask Training},
  author = {Yan, Yibin and Xu, Jilan and Di, Shangzhe and Liu, Yikun and Shi, Yudi and Chen, Qirui and Li, Zeqian and Huang, Yifei and Xie, Weidi},
  booktitle = {Proceedings of the IEEE/CVF International Conference on Computer Vision (ICCV)},
  year = {2025},
  pages = {9900--9912},
  month = {October},
  url = {https://api.semanticscholar.org/CorpusID:278165314}
}

@inproceedings{Chen2025LiveLV,
  title = {LiveCC: Learning Video LLM with Streaming Speech Transcription at Scale},
  author = {Chen, Joya and Zeng, Ziyun and Lin, Yiqi and Li, Wei and Ma, Zejun and Shou, Mike Zheng},
  booktitle = {Proceedings of the IEEE/CVF Conference on Computer Vision and Pattern Recognition (CVPR)},
  year = {2025},
  pages = {29083--29095},
  month = {June},
  url = {https://api.semanticscholar.org/CorpusID:277993668}
}

@inproceedings{Ding2025StreamMindUF,
  title = {StreamMind: Unlocking Full Frame Rate Streaming Video Dialogue through Event-Gated Cognition},
  author = {Ding, Xin and Wu, Hao and Yang, Yifan and Jiang, Shiqi and Zhang, Qianxi and Bai, Donglin and Chen, Zhibo and Cao, Ting},
  booktitle = {Proceedings of the IEEE/CVF International Conference on Computer Vision (ICCV)},
  year = {2025},
  pages = {13448--13459},
  month = {October},
  url = {https://api.semanticscholar.org/CorpusID:276903154}
}

@inproceedings{Fu2025ViSpeakVI,
  title = {ViSpeak: Visual Instruction Feedback in Streaming Videos},
  author = {Fu, Shenghao and Yang, Qize and Li, Yuan-Ming and Peng, Yi-Xing and Lin, Kun-Yu and Wei, Xihan and Hu, Jian-Fang and Xie, Xiaohua and Zheng, Wei-Shi},
  booktitle = {Proceedings of the IEEE/CVF International Conference on Computer Vision (ICCV)},
  year = {2025},
  pages = {21778--21788},
  month = {October},
  url = {https://api.semanticscholar.org/CorpusID:277066592}
}

@inproceedings{Azad2026StreamReadyLW,
  title = {StreamReady: Learning What to Answer and When in Long Streaming Videos},
  author = {Azad, Shehreen and Vineet, Vibhav and Rawat, Yogesh S},
  booktitle = {Proceedings of the IEEE/CVF Conference on Computer Vision and Pattern Recognition (CVPR)},
  year = {2026},
  pages = {40494--40504},
  month = {June},
  url = {https://api.semanticscholar.org/CorpusID:286377314}
}

@inproceedings{Xie2026StreamRAGER,
  title = {StreamRAG: Enhancing Real-Time Video Understanding with Retrieval Augmentation},
  author = {Xie, Junlin and Zheng, Quanlong and Zhang, Ruifei and Wang, Kuo and Zhang, Yanhao and Luo, Jinguo and Lu, Haonan and Wan, Xiang and Li, Guanbin},
  booktitle = {Proceedings of the IEEE/CVF Conference on Computer Vision and Pattern Recognition (CVPR)},
  year = {2026},
  pages = {38870--38879},
  month = {June}
}

@inproceedings{Chen2026ScalingTL,
  title = {Scaling the Long Video Understanding of Multimodal Large Language Models via Visual Memory Mechanism},
  author = {Chen, Tao and Zhang, Kun and Wu, Qiong and Chen, Xiao and Chang, Chao and Sun, Xiaoshuai and Zhou, Yiyi and Ji, Rongrong},
  booktitle = {Proceedings of the IEEE/CVF Conference on Computer Vision and Pattern Recognition (CVPR)},
  year = {2026},
  pages = {31877--31888},
  month = {June},
  url = {https://api.semanticscholar.org/CorpusID:286977474}
}

@article{Ge2026WhatSA,
  title = {What Should a Streaming Video Model Remember?},
  author = {Ge, Haonan and Wang, Yiwei and Wu, Hang and Cai, Yujun},
  journal = {ArXiv},
  year = {2026},
  volume = {abs/2606.16353},
  url = {https://api.semanticscholar.org/CorpusID:289299850}
}

@article{Wu2026SemanticAwareAV,
  title = {Semantic-Aware Adaptive Visual Memory for Streaming Video Understanding},
  author = {Wu, Hang and Mathews, Sherin Mary and Cai, Yujun and Yang, Ming-Hsuan and Wang, Yiwei},
  journal = {ArXiv},
  year = {2026},
  volume = {abs/2605.07897},
  url = {https://api.semanticscholar.org/CorpusID:288147794}
}

@inproceedings{Wang2025AcceleratingSV,
  title = {Accelerating Streaming Video Large Language Models via Hierarchical Token Compression},
  author = {Wang, Yiyu and Liu, Xuyang and Gui, Xiyan and Lin, Xinying and Yang, Boxue and Liao, Chenfei and Chen, Tailai and Zhang, Linfeng},
  booktitle = {Proceedings of the IEEE/CVF Conference on Computer Vision and Pattern Recognition (CVPR)},
  year = {2026},
  pages = {18523--18533},
  month = {June},
  url = {https://api.semanticscholar.org/CorpusID:283450494}
}

@inproceedings{Wang2024VideoTreeAT,
  title = {VideoTree: Adaptive Tree-based Video Representation for LLM Reasoning on Long Videos},
  author = {Wang, Ziyang and Yu, Shoubin and Stengel-Eskin, Elias and Yoon, Jaehong and Cheng, Feng and Bertasius, Gedas and Bansal, Mohit},
  booktitle = {Proceedings of the IEEE/CVF Conference on Computer Vision and Pattern Recognition (CVPR)},
  year = {2025},
  pages = {3272--3283},
  month = {June},
  url = {https://api.semanticscholar.org/CorpusID:270094803}
}

@inproceedings{Yeo2025WorldMMDM,
  title = {WorldMM: Dynamic Multimodal Memory Agent for Long Video Reasoning},
  author = {Yeo, Woongyeong and Kim, Kangsan and Yoon, Jaehong and Hwang, Sung Ju},
  booktitle = {Proceedings of the IEEE/CVF Conference on Computer Vision and Pattern Recognition (CVPR)},
  year = {2026},
  pages = {25599--25609},
  month = {June},
  url = {https://api.semanticscholar.org/CorpusID:283458398}
}

@inproceedings{Qiu2026LongVideoR1SN,
  title = {LongVideo-R1: Smart Navigation for Low-cost Long Video Understanding},
  author = {Qiu, Jihao and Xie, Lingxi and Huo, Xinyue and Tian, Qi and Ye, Qixiang},
  booktitle = {Proceedings of the IEEE/CVF Conference on Computer Vision and Pattern Recognition (CVPR)},
  year = {2026},
  pages = {40505--40515},
  month = {June},
  url = {https://api.semanticscholar.org/CorpusID:286000722}
}

@inproceedings{Xun2025RTVBenchBM,
  title = {RTV-Bench: Benchmarking MLLM Continuous Perception, Understanding and Reasoning through Real-Time Video},
  author = {Xun, ShuHang and Tao, Sicheng and Li, Jungang and Shi, Yibo and Lin, Zhixin and Zhu, Zhanhui and Yan, Yibo and Li, Hanqian and Zhang, LingHao and Wang, Shikang and Liu, Yixin and Zhang, Hanbo and Ma, Ying and Hu, Xuming},
  booktitle = {Advances in Neural Information Processing Systems},
  year = {2025},
  volume = {38},
  publisher = {Curran Associates, Inc.},
  doi = {10.52202/085713-0600},
  url = {https://api.semanticscholar.org/CorpusID:278326974}
}

@article{Hao2026DeltaKVRK,
  title = {DeltaKV: Residual-Based KV Cache Compression via Long-Range Similarity},
  author = {Hao, Jitai and Huang, Qiang and Wang, Yaowei and Zhang, Min and Yu, Jun},
  journal = {ArXiv},
  year = {2026},
  volume = {abs/2602.08005},
  url = {https://api.semanticscholar.org/CorpusID:285454299}
}

@inproceedings{Hao2025ATI,
  title = {A Token is Worth over 1,000 Tokens: Efficient Knowledge Distillation through Low-Rank Clone},
  author = {Hao, Jitai and Huang, Qiang and Liu, Hao and Xiao, Xinyan and Ren, Zhaochun and Yu, Jun},
  booktitle = {Advances in Neural Information Processing Systems},
  year = {2025},
  volume = {38},
  pages = {53678--53710},
  publisher = {Curran Associates, Inc.},
  doi = {10.52202/085713-1791},
  url = {https://api.semanticscholar.org/CorpusID:278740403}
}

@inproceedings{Hao2025UniXMM,
  title = {Uni-X: Mitigating Modality Conflict with a Two-End-Separated Architecture for Unified Multimodal Models},
  author = {Hao, Jitai and Liu, Hao and Xiao, Xinyan and Huang, Qiang and Yu, Jun},
  booktitle = {International Conference on Learning Representations},
  year = {2026},
  volume = {2026},
  pages = {123737--123758},
  url = {https://api.semanticscholar.org/CorpusID:281674376}
}

@inproceedings{Alvar2025DivPruneDV,
  title = {DivPrune: Diversity-based Visual Token Pruning for Large Multimodal Models},
  author = {Alvar, Saeed Ranjbar and Singh, Gursimran and Akbari, Mohammad and Zhang, Yong},
  booktitle = {Proceedings of the IEEE/CVF Conference on Computer Vision and Pattern Recognition (CVPR)},
  year = {2025},
  pages = {9392--9401},
  month = {June},
  url = {https://api.semanticscholar.org/CorpusID:276775957}
}

\appendix

\section{Other Related Work}
\label{appendix:other_related_work}

\paragraph{Model Architecture and Adaptation.}
StreamFormer~\citep{Yan2025LearningSV}, LiveCC~\citep{Chen2025LiveLV}, and Streamo~\citep{Xia2025StreamingVI} improve streaming representations through architectural adaptation and training, while Uni-X~\citep{Hao2025UniXMM} addresses modality conflict with a two-end-separated architecture.
ShallowStream instead retains the pretrained MLLM and schedules computation across its depth, using shallow layers for continuous indexing and the full model for query-time answering.

\paragraph{Proactive Interaction and Routing.}
StreamMind~\citep{Ding2025StreamMindUF}, Dispider~\citep{Qian2025DispiderEV}, and StreamBridge~\citep{Wang2025StreamBridgeTY} use event detection, decision modules, or activation models to determine when to invoke language-model inference or generate proactive responses; ViSpeak~\citep{Fu2025ViSpeakVI} and StreamReady~\citep{Azad2026StreamReadyLW} additionally consider visual interaction cues and evidence readiness.
ShallowStream addresses a distinct routing problem: after a question arrives, its text-only Gate determines whether that question requires historical retrieval.

\paragraph{Memory Construction and Retrieval.}
StreamRAG~\citep{Xie2026StreamRAGER} and EventMemAgent~\citep{Wen2026EventMemAgentHE} organize event-level evidence; MuKV~\citep{Xiao2026MuKVMK}, FlexMem~\citep{Chen2026ScalingTL}, and DeltaKV~\citep{Hao2026DeltaKVRK} explore multiscale KV retrieval, compressed visual states, and residual KV representations, respectively.
Concurrent SelectStream and SAVEMem further investigate budgeted historical memory and query-adaptive recall~\citep{Ge2026WhatSA, Wu2026SemanticAwareAV}.
These methods vary how history is represented, compressed, or retrieved.
ShallowStream instead focuses on the cost of constructing that history: it builds a shallow index, optionally compresses older states into clusters, and selects evidence before full-depth visual prefill.

\paragraph{Efficient Streaming Execution.}
Streaming Token Compression combines visual-feature reuse with token pruning~\citep{Wang2025AcceleratingSV}, while TimeChat-Online and StreamingTOM compress visual tokens~\citep{Yao2025TimeChatOnline8V, Chen2025StreamingTOMST}; Low-Rank Clone improves knowledge-transfer efficiency through distillation~\citep{Hao2025ATI}.
ShallowStream complements these approaches by reducing Transformer depth during continuous stream processing and invoking the full model only for selected query-time evidence, without task-specific training.

\paragraph{Query-Driven Long-Video Reasoning.}
VideoTree, WorldMM, and LongVideo-R1 use hierarchical or multi-memory search to acquire question-relevant evidence~\citep{Wang2024VideoTreeAT, Yeo2025WorldMMDM, Qiu2026LongVideoR1SN}.
ShallowStream similarly performs query-conditioned evidence selection, but builds its index causally before future questions and frames are available.

\paragraph{Streaming and Interactive Evaluation.}
RTV-Bench evaluates evolving answers, ProReady-QA examines evidence timing and proactive questions, and ViSpeak-Bench tests visually triggered interaction~\citep{Xun2025RTVBenchBM, Azad2026StreamReadyLW, Fu2025ViSpeakVI}.
Together with OVO-Bench, StreamingBench, and SVBench~\citep{Li2025OVOBenchHF, Lin2024StreamingBenchAT, Yang2025SVBenchAB}, these benchmarks cover complementary aspects of streaming understanding.
Our main experiments use OVO-Bench and StreamingBench to evaluate retrospective evidence tracing and real-time visual understanding under causal access to the observed video prefix.

\section{Attention Context and Retrieval Depth}
\label{appendix:sliding_window_retrieval_depth}

The layer-wise diagnostic in Figure~\ref{fig:layer_retrieval_ability} adopts ReKV's stream-time prefill configuration~\citep{Di2025StreamingVQ}, in which each incoming video unit attends to an attention sink and a bounded local window.
This constraint limits stream-time attention but not retrieval: the full-history shallow index covers the entire observed stream, enabling effective shallow video--question matching without deeper prefill.

A dense complete-causal-prefix control further shows that the preferred retrieval depth depends on the attention regime.
Full-history attention can improve retrieval representations at deeper layers, but its prefill cost grows with the accumulated history, preventing bounded per-unit computation on open-ended streams.
ShallowStream therefore targets bounded-context streaming, where retrieval signals emerge early and full-depth computation is reserved for query-relevant evidence.

\begin{table}[!t]
\centering
\small
\renewcommand{\arraystretch}{1.3}
\caption{\textbf{Query-logit Gate calibration on the benchmark-independent query-only set.}}
\label{tab:query_gate_calibration}
\begin{tabular}{lcccc}
\toprule
\textbf{Backbone} & $\boldsymbol{\tau_g}$ & \textbf{Precision} & \textbf{95\% lower bound} & \textbf{Recall} \\
\midrule
\textbf{Qwen3-VL-8B}        & 11.0  & 96.04\% & 91.47\% & 97.00\% \\
\textbf{LLaVA-OneVision-7B} & 0.875 & 97.78\% & 90.63\% & 44.00\% \\
\bottomrule
\end{tabular}
\end{table}

\section{Query-Logit Gate Calibration}
\label{appendix:query_gate_calibration}

We construct a benchmark-independent, query-only calibration set of synthetically generated questions labeled using the same retrospective-versus-recent definition as the ten-shot prompt; no OVO-Bench or StreamingBench queries are included.
For each backbone, we apply the prompt and compute $s(q)=\ell_{\mathrm{ret}}(q)-\ell_{\mathrm{recent}}(q)$.
We then select a backbone-specific threshold using a precision-first criterion requiring the one-sided 95\% Wilson lower bound on retrieval precision to reach at least 90\%, and freeze it before evaluating either benchmark (Table \ref{tab:query_gate_calibration}).

\begin{table}[!t]
\centering
\small
\renewcommand{\arraystretch}{1.3}
\caption{\textbf{Main implementation settings for the two evaluated backbones.}}
\label{tab:implementation_settings}
\begin{tabular}{lrr}
\toprule
\textbf{Setting} & \textbf{Qwen3-VL-8B} & \textbf{LLaVA-OneVision-7B} \\
\midrule
\rowcolor{black!6} 
\multicolumn{3}{l}{\textit{Backbone and Stream Processing}} \\
\textbf{Pruning boundary $\boldsymbol{P}$} & 5 & 4 \\
\textbf{Sampling rate (OVO / StreamingBench)} & 1 / 1 FPS & 1 / 1 FPS \\
\textbf{Video unit} & 2 frames & 1 frame \\
\textbf{Stream prefill window} & 64 units, 128 frames & 128 units, 128 frames \\
\rowcolor{black!6} 
\multicolumn{3}{l}{\textit{Query Routing}} \\
\textbf{Gate prompt} & ten shot & ten shot \\
\textbf{Shared Gate calibration set} & 200 query-only examples & 200 query-only examples \\
\textbf{Gate threshold $\boldsymbol{\tau_g}$} & 11.0 & 0.875 \\
\rowcolor{black!6} 
\multicolumn{3}{l}{\textit{Evidence Selection and Generation}} \\
\textbf{Query representation} & last prompt token & last prompt token \\
\textbf{Visual candidates per shallow layer} & 64 tokens & 64 tokens \\
\textbf{Vote-ranked candidate pool} & 32 historical units & 32 historical units \\
\textbf{Selected historical evidence} & 8 units & 8 units \\
\textbf{Diversity selection} & max-min & max-min \\
\textbf{Temporal expansion per selected unit} & 1 preceding unit & 1 preceding unit \\
\textbf{Recent context, retrieval / recent-only} & 6 / 2 units & 6 / 2 units \\
\textbf{Maximum new tokens} & 16 & 16 \\
\bottomrule
\end{tabular}
\end{table}

\section{Implementation Details}
\label{appendix:implementation_settings}

Table~\ref{tab:implementation_settings} summarizes the fixed configurations used in our main experiments. Both backbones use 1 FPS on both benchmarks and share the same evidence selection and decoding budgets, while the pruning boundary, Gate threshold, and video unit definition follow their respective architectures. Shallow KVs and input-level visual states are archived in host memory. In LC-off runs, this archive preserves every detailed unit; in LC-on runs, units outside the detailed window are replaced by fixed-size cluster KVs and one input representative per cluster. Query latency is measured from query entry to generation completion and therefore includes host-to-device transfer and selected-context assembly.

\begin{table}[!t]
\centering
\small
\renewcommand{\arraystretch}{1.3}
\caption{\textbf{Cost summary under the final calibrated Gate and token-vote retriever.} Stream-time prefill is averaged over the same five long videos; query components use the matched full-history LC-off setting.}
\label{tab:pruning_boundary_cost}
\begin{tabular}{llr}
\toprule
\textbf{Measurement} & \textbf{Setting or Component} & \textbf{Cost} \\
\midrule
\multirow{3}{*}{\textbf{Stream-time prefill}}
  & $P=1$ & $9.47$ ms/frame \\
  & $P=5$ & $10.72$ ms/frame \\
  & $P=19$ & $15.53$ ms/frame \\
\midrule
\multirow{3}{*}{\textbf{Query-time computation}}
  & Full query & $1.759$ s/query \\
  & Gate & $92.7$ ms/query \\
  & Evidence selection & $106.0$ ms/query \\
\bottomrule
\end{tabular}
\end{table}

\section{Pruning-Boundary Accuracy and Cost}
\label{appendix:pruning_boundary_sensitivity}

\begin{wrapfigure}[16]{r}{0.6\textwidth}
  \vspace{-1.0em}
  \centering
  \includegraphics[width=0.99\linewidth]{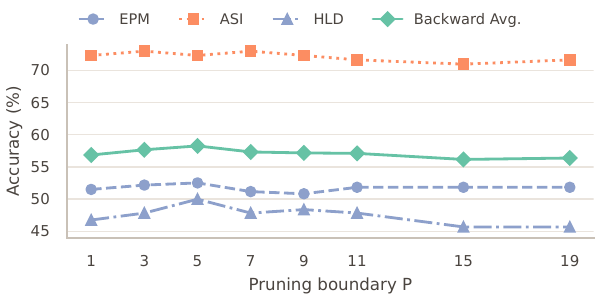}
  \vspace{-0.5em}
  \caption{\textbf{Sensitivity to the pruning boundary $P$ on OVO-Bench Backward under the final calibrated Gate and token-vote retriever.} Routing and evidence selection are held fixed across depths.}
  \label{fig:ovobench-prune-layer-sensitivity}
\end{wrapfigure}

The first five layers already form an effective visual evidence index, allowing ShallowStream to retain its strongest accuracy without unnecessary continuous-stream depth. Figure~\ref{fig:ovobench-prune-layer-sensitivity} reports a full OVO-Bench Backward sweep that varies only the pruning boundary while holding routing, evidence selection, and recent-only outputs fixed. The $P=5$ boundary, selected beforehand from LVBench, achieves the strongest EPM, HLD, and Backward average, with the latter reaching $58.27$; increasing the boundary through $P=19$ provides no further gain.

\paragraph{Shallow Processing Preserves Accuracy at Lower Cost.}
ShallowStream’s efficiency stems from applying shallow computation to every incoming frame while reserving deeper processing for sparse queries. Because stream-time prefill is repeated continuously, even modest per-frame savings accumulate substantially over long videos.
As shown in Table~\ref{tab:pruning_boundary_cost}, the observation-guided boundary $P=5$ reduces per-frame prefill by 31.0\% relative to $P=19$, while achieving the highest accuracy in Figure~\ref{fig:ovobench-prune-layer-sensitivity}.
This result indicates that deeper continuous prefill adds recurring cost without improving retrieval quality.
At query time, Gate routing and evidence selection contribute only a small fraction of the total computation, so their overhead does not offset the savings from shallow encoding.
ShallowStream therefore maintains a favorable accuracy-efficiency trade-off even with its stronger query-time evidence selection.

\begin{figure}[ht]
  \centering
  \includegraphics[width=0.99\linewidth]{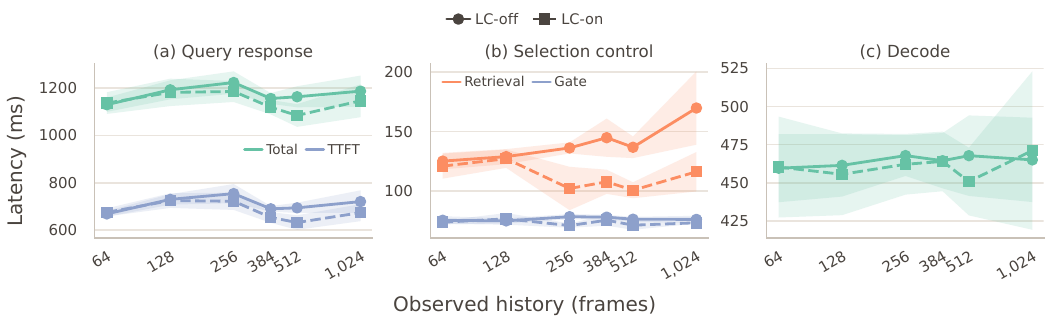}
  \caption{\textbf{Query-time scaling with retained history on an NVIDIA RTX 5090.}
  Results span six history lengths from 64 to 1,024 frames. \emph{(a)} Total query latency and time to first token (TTFT). \emph{(b)} Evidence-selection and Gate latency. \emph{(c)} Decoding latency.
  Solid circles and dashed squares denote full-history LC-off and the matched final LC-on setting, respectively. Curves show means over five retrieval queries; shaded regions indicate 95\% confidence intervals.}
  \label{fig:appendix-query-history-scaling}
\end{figure}

\section{Query-Time Scaling with Retained History}
\label{appendix:query_history_scaling}

To assess whether query cost grows with retained history, we compare full-history LC-off with the matched final LC-on setting across a $16\times$ range of history lengths (Figure \ref{fig:appendix-query-history-scaling}).
Wall-clock latency includes transfers from the host-resident archive and selected-context assembly.
Under both settings, total query latency and time to first token remain nearly constant as history grows, while Gate and decoding costs are largely invariant.
Long-cluster compression also keeps evidence-selection cost essentially constant, while Gate and decoding latency remain nearly invariant. Thus, even the complete shallow history remains responsive, while compression additionally bounds the history-dependent selection stage.

\section{Ten-Shot Router Prompt}
\label{appendix:ten_shot_router_prompt}

For reproducibility, the complete prompt template used by our ten-shot
query-logit router is given below. \texttt{[Input video question]} denotes the
question inserted at inference time.

\begin{tcblisting}{
  listing only,
  breakable,
  colback=gray!3,
  colframe=black!35,
  boxrule=0.5pt,
  arc=1mm,
  left=1mm,
  right=1mm,
  top=0.28453em,
  bottom=0.28453em,
  title={Ten-Shot Query-Logit Router Prompt},
  coltitle=black,
  fonttitle=\small\bfseries,
  listing options={
    basicstyle=\ttfamily\scriptsize,
    breaklines=true,
    columns=fullflexible,
    keepspaces=true,
    showstringspaces=false,
    aboveskip=0em,
    belowskip=0em
  }
}
You are deciding whether answering a video question requires retrieving older video memory. The system already has the latest video segment immediately available. Do not answer the video question itself.

Routing options:
A. Answering reliably requires inspecting earlier video memory.
B. The latest video segment is sufficient; older video memory is unnecessary.

Here are labeled routing examples:

Example 1:
Video question: What object is the person holding in the latest visible moment?
Correct routing option: B

Example 2:
Video question: Where did the person leave the keys near the beginning of the video?
Correct routing option: A

Example 3:
Video question: What word is currently displayed on the screen?
Correct routing option: B

Example 4:
Video question: How many times did the bell ring over the course of the video?
Correct routing option: A

Example 5:
Video question: Which item did the cook pick up first?
Correct routing option: A

Example 6:
Video question: What color is the vehicle visible in the latest scene?
Correct routing option: B

Example 7:
Video question: What was inside the box before it was emptied?
Correct routing option: A

Example 8:
Video question: What action is the person performing right now?
Correct routing option: B

Example 9:
Video question: Which object is immediately to the left of the cup in the latest frame?
Correct routing option: B

Example 10:
Video question: How has the room changed compared with its appearance earlier in the video?
Correct routing option: A

Now route this video question:
<video_question>
[Input video question]
</video_question>

Output only A or B.
Routing answer:
\end{tcblisting}

\end{document}